\documentclass[journal]{IEEEtran}
\ifCLASSINFOpdf
  \usepackage[pdftex]{graphicx}
\else
\fi

\usepackage{mathrsfs}
\usepackage{multirow}
\usepackage{booktabs}
\usepackage{hyperref}
\usepackage{color}
\usepackage{amsmath, bm}
\usepackage{algorithm}
\usepackage{algorithmic}
\usepackage{tabu}
\usepackage{cancel}
\usepackage{subfig}
\usepackage[table]{xcolor}
\usepackage{fp}  

\usepackage{flushend}

\usepackage{mathrsfs}

\ifCLASSINFOpdf
\else
\fi
\usepackage{amssymb}
\usepackage{mathrsfs}
\usepackage{amsmath}
\usepackage{bm}
\usepackage{makecell}
\usepackage{multirow}
\usepackage{colortbl}
\usepackage{soul}
\usepackage{caption}
\usepackage[figuresleft]{rotating}
\definecolor{chromeyellow}{rgb}{1.0, 0.65, 0.0}
\definecolor{carmine}{rgb}{0.59, 0.0, 0.09}
\usepackage[percent]{overpic}

\begin{document}

\captionsetup[table]{skip=0pt}
%
\title{MaeFuse: Transferring Omni Features with Pretrained Masked Autoencoders for Infrared and Visible Image Fusion via Guided Training}
%
%
%

\author{Jiayang Li,
        Junjun~Jiang,~\IEEEmembership{Senior Member,~IEEE,~}
        Pengwei Liang,
        Jiayi Ma,~\IEEEmembership{Senior Member,~IEEE,~}
        Liqiang Nie,~\IEEEmembership{Senior Member,~IEEE}
 

\IEEEcompsocthanksitem 

\thanks{J. Li, J. Jiang and P. Liang are with the School of Computer Science and Technology, Harbin Institute of Technology, Harbin 150001. E-mail: lijiayang.cs@gmail.com, jiangjunjun@hit.edu.cn, erfect2020@gmail.com.}
\thanks{J. Ma is with the Electronic Information School, Wuhan University, Wuhan 430072, China. E-mail: jyma2010@gmail.com.}
\thanks{L. Nie is with the School of Computer Science and Technology, Harbin Institute of Technology (Shenzhen), Shenzhen 518055, China. E-mail: nieliqiang@gmail.com.}
}

\markboth{Journal of \LaTeX\ Class Files,~Vol.~14, No.~8, August~2021}%
{Shell \MakeLowercase{\textit{\emph{\emph{et al. } }}}: Bare Demo of IEEEtran.cls for IEEE Journals}
%



\maketitle

\begin{abstract}
In this paper, we introduce MaeFuse, a novel autoencoder model designed for Infrared and Visible
Image Fusion (IVIF). The existing approaches for
image fusion often rely on training combined with
downstream tasks to obtain high-level visual information, which is effective in emphasizing target
objects and delivering impressive results in visual
quality and task-specific applications. Instead of being driven by downstream tasks, our model called MaeFuse utilizes
a pretrained encoder from Masked Autoencoders
(MAE), which facilities the omni features extraction for low-level reconstruction and high-level vision tasks, to obtain perception friendly features
with a low cost. In order to eliminate the domain
gap of different modal features and the block effect
caused by the MAE encoder, we further develop a
guided training strategy. This strategy is meticulously crafted to ensure that the fusion layer seamlessly adjusts to the feature space of the encoder,
gradually enhancing the fusion performance. The proposed method can facilitate
the comprehensive integration of feature vectors
from both infrared and visible modalities, thus preserving the rich details inherent in each modal. MaeFuse not
only introduces a novel perspective in the realm of
fusion techniques but also stands out with impressive performance across various public datasets. The code is available at \href{https://github.com/Henry-Lee-real/MaeFuse}{https://github.com/Henry-Lee-real/MaeFuse}.
\end{abstract}

\begin{IEEEkeywords}
Image fusion, Vision Transformer, Masked Autoencoder, Guided training
\end{IEEEkeywords}

%
\IEEEpeerreviewmaketitle

\section{Introduction}\label{sec1}


\IEEEPARstart{M}{ultimodal} sensing technology has significantly contributed to the widespread use of multimodal imaging in various fields, and the fusion of infrared and visible image is the most common application technique. Infrared and Visible Image Fusion (IVIF) can merge the complementary information of these two modalities. Infrared images display thermal contours and are unaffected by lighting effects, but lacks texture and color accuracy. Visible images can capture texture but depend on lighting. Fused images can combine their strengths, and thus enhancing visual clarity and the effectiveness of downstream visual tasks~\cite{RTFNet,VIFB} and surpassing the results achievable by each modality when used separately. 

To effectively fuse the useful elements of both modalities, diverse fusion techniques and training approaches have been developed. Traditional image fusion methods focus on enhancing visual effects by implementing various strategies, including subspace transforms~\cite{fu2016infrared}, multi-scale transforms~\cite{chen2020infrared}, sparse representations~\cite{li2020mdlatlrr}, and saliency analysis~\cite{ma2017infrared}. With the rise of deep learning, researchers have crafted deep learning models to enhance the visual quality of fusion outcomes. The design of architectures is primarily divided into two categories: directly concat infrared and visible images for fusion (pre-fusion), and first extract infrared and visible features separately before fusion (post-fusion). In the first approach, there are implementations utilizing residual connections~\cite{RFN-Nest} and adversarial learning techniques~\cite{FusionGAN,DDcGAN}. In the second approach, some directly employ pretrained convolutional encoders~\cite{DenseFuse}, some integrate fusion in different frequency domains~\cite{DIDFuse}, some use Transformers as the backbone network~\cite{SwinFusion,wang2022swinfuse}, and there are also some methods combining Transformers with convolutional architectures~\cite{CDDFuse}. These approaches have retained some degree of visual texture information. However, the fused outcomes often display a particular style, such as increased brightness or prominent gradient features. Additionally, in certain situations, they lack the ability to make adaptive selections and adjustments, such as dimming illumination in specific areas. Fundamentally, object information, also referred to as high-level visual information, is not adequately captured and is therefore overlooked during fusion. This insufficient capture of high-level semantic cues can lead to ambiguities in identifying visual targets and potentially result in suboptimal performance in downstream tasks.

\begin{figure}[t] 
  \centering 
  \includegraphics[width=\columnwidth]{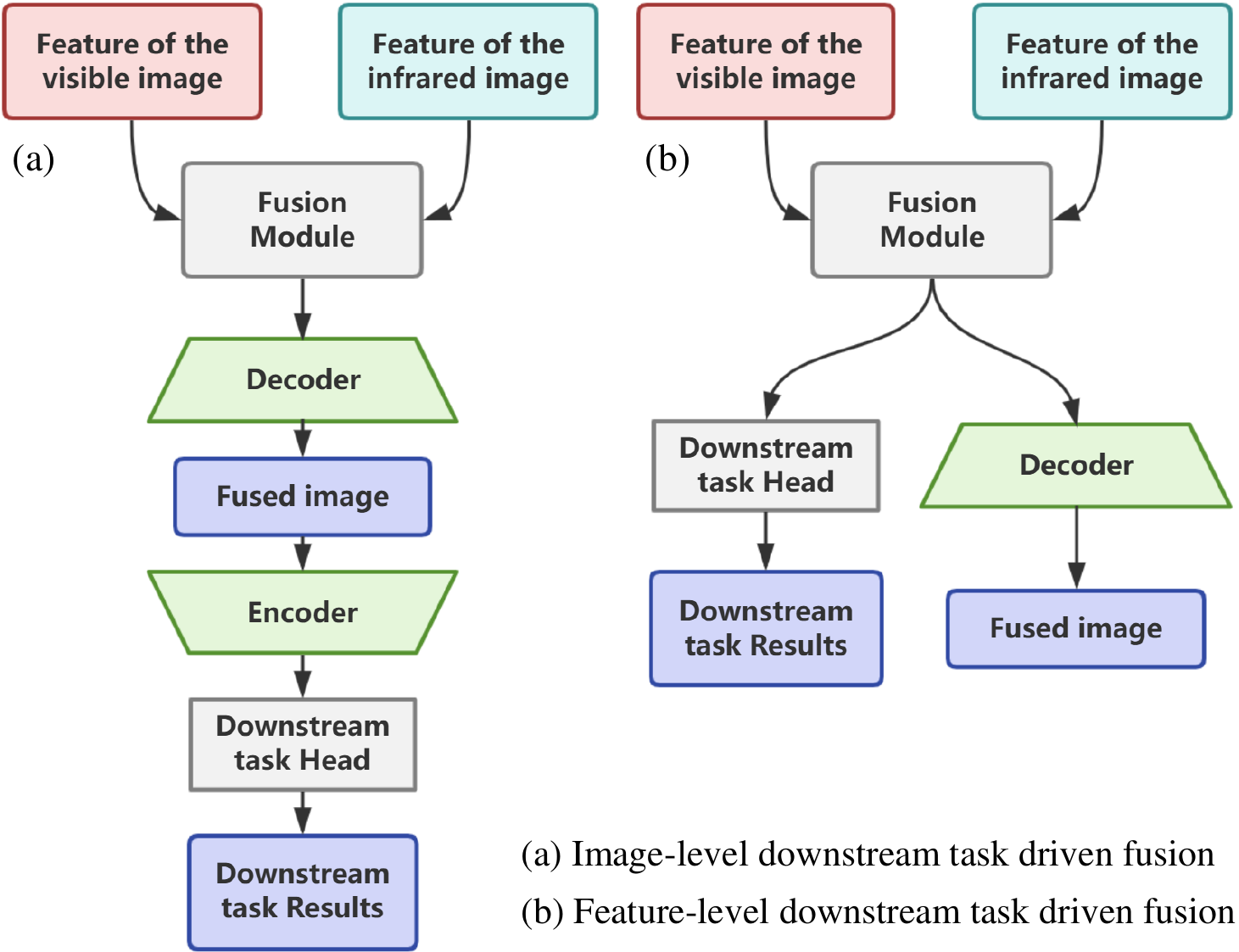} 
  \caption{Diagram illustrating image-level downstream task-driven fusion and feature-level downstream task-driven fusion.}
  \vspace{-15pt}
  \label{fig:2_me}
\end{figure}

In an effort to learn high-level semantic information, researchers have employed downstream tasks to drive fusion networks to better learn the high-level semantic information of target objects. Some have utilized semantic segmentation tasks~\cite{SeAfusion,Metafusion}, while others have employed object detection~\cite{TAL,BDLFusion}. However, in the aforementioned methods, the fusion modules and downstream task modules within the network architecture do not undergo effective joint learning. This is because these methods typically feed the fused images into the downstream task models for processing. During the gradient backpropagation process, gradients first flow from the downstream task network, then through the decoder, and finally reach the fusion network. This sequence causes the gradient information to deviate to some extent. Additionally, training directly on limited annotated data makes it challenging to integrate high-level visual information into the fusion module through downstream tasks. Therefore, researchers proposed directly using the fused features as the features extracted by the downstream task network for subsequent operations~\cite{PSfusion}. This approach allows for more accurate gradient backpropagation and better performance, thereby enhancing the encoder's ability to extract high-level visual information under limited data conditions. Fig.~\ref{fig:2_me} provides a more intuitive illustration of the differences between the two aforementioned architectures. There are also suggestions to utilize meta-learning training methods~\cite{Metafusion}, allowing the model to progressively learn relevant object information and fully utilize the information brought by downstream tasks. However, these methods still introduced complexities in model architecture and training approaches, and still face the issue of insufficient labeled data, leading to a risk of overfitting on training data.

Based on previous methods, the issues can generally be summarized as follows: 1) There is a deficiency of extensive paired datasets with labels for downstream tasks; 2) The need to inject high-level semantic information into the model to enhance both the visual fusion effects and performance of downstream tasks. To address the issue of scarce data, we employed pretrained models to extract visual features, as these models possess rich visual priors that can effectively capture key visual information. Furthermore, to inject high-level semantic information, we decided to use pretrained networks that already contain high-level semantic information, allowing us to naturally preserve high-level semantic information within the visual features.

\begin{figure}[h] 
  \centering 
  \includegraphics[width=\columnwidth]{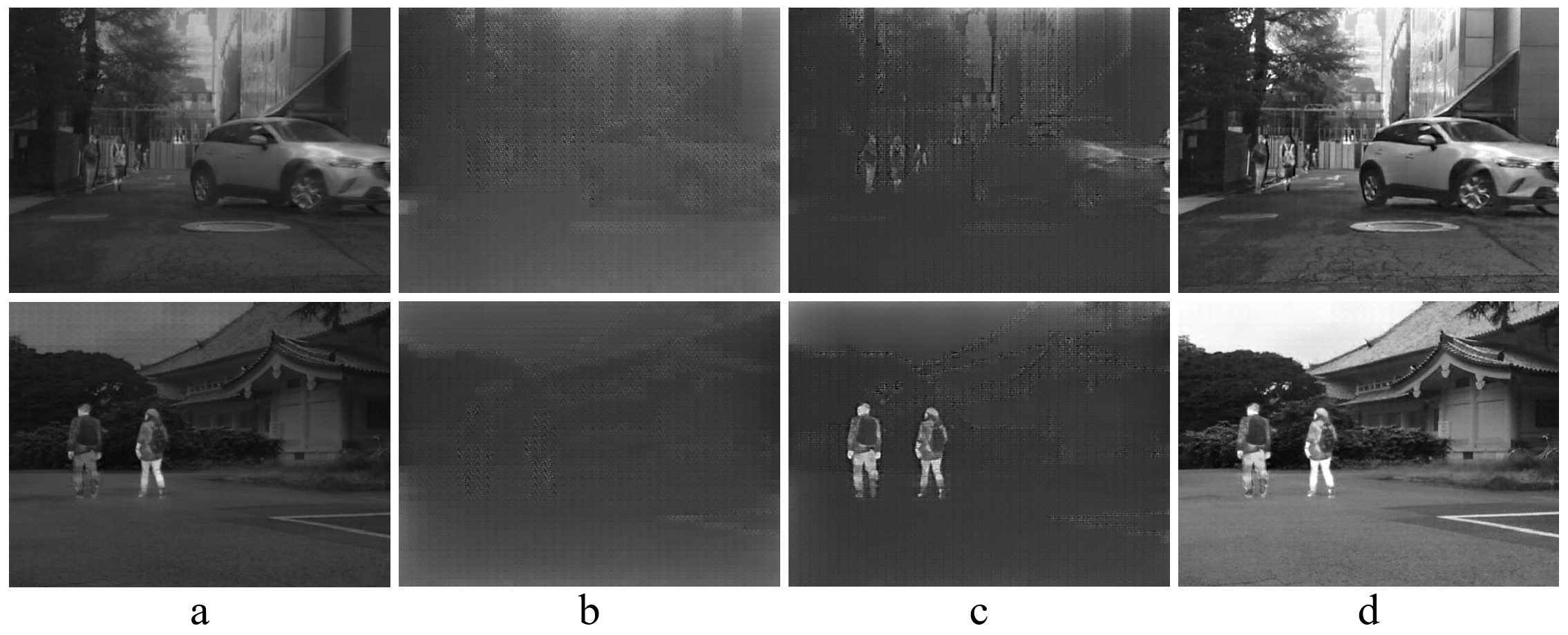} 
  \caption{The fusion results with different fusion strategies: (a) mean fusion for two features, (b) cross-attention based fusion without alignment in the feature domain, (c) cross-attention based fusion with alignment in the feature domain, and (d) our MaeFuse. Here we only show the grayscale images for better comparison.}
  \vspace{-15pt}
  \label{fig:intro}
\end{figure}

With the development of deep learning and pretraining models, a plethora of visual pretraining models such as MAE~\cite{Mae}, Diffusion Model~\cite{ho2020denoising}, and Vision Mamba~\cite{zhu2024vision} have emerged recently. For the Diffusion Model, although there are optimizations based on the Expectation-Maximization algorithm like DDFM~\cite{zhao2023ddfm} and prompt-based approaches like Diff-IF~\cite{yi2024diff}, they face practical limitations due to speed issues associated with sampling. Moreover, the images restored through sampling exhibit significant deviations from real images, such as distorted text information. As for Vision Mamba based methods~\cite{li2024mambadfuse,ma2024s4fusion,peng2024fusionmamba}, it is unclear whether this model can simultaneously capture high-level semantic and low-level texture information. MAE~\cite{Mae}, an autoregressive pretrained model, effectively extracts both high-level~\cite{mae-dect,liu2023integrally} and low-level~\cite{liang2024image} visual information, making it an ideal choice for robust feature extraction to achieve better fusion. In this paper, we propose using the pretrained MAE encoder to extract features from infrared and visible images for better fusion. Because MAE is trained via self-supervised learning, it has a strong ability to extract both high-level and low-level visual information. As a result, the fusion outcomes can maintain low-level texture and color information while also preserving robust high-level object information. We refer to these visual features that cover both high-level and low-level aspects as 'omni features'. By employing a pretrained model with strong representational capabilities, we can expect to reduce the dependency on data in the IVIF task, where paired samples with downstream task labels are typically scarce.

However, as shown in Fig.~\ref{fig:intro}, fusing infrared and visible features using standard cross-attention~\cite{CrossFuse,SegMif}, may result in the loss of a substantial amount of detailed information. Consequently, the fused images exhibit obvious block effects and inharmonious regions. The block effect refers to the phenomenon of discontinuity in color and texture between patches due to the ViT architecture dividing the image into patches for learning. To this end, in this paper we further introduce a guided training strategy to mitigate this issue. This strategy is employed to expedite the alignment of the fusion layer output with the encoder feature domain, thereby avoiding the potential risk of converging to local optima. By adopting this approach, a simple yet effective fusion network is developed, which successfully extracts and retains both low-level image reconstruction and high-level visual features. Our contributions can be distilled into two main aspects:

\begin{itemize}
   \item Employing a pretrained encoder, \emph{i.e.}, MAE, as the encoder for fusion tasks enables the fusion network to preserve comprehensive low-level and high-level visual information. This approach addresses the issue of lacking high-level visual information in fusion features and also simplifies the overall network structure. Additionally, it also alleviate the problem of insufficient labeled data during training.
   \item A guided training strategy is proposed, aiding the fusion layer in rapidly adapting to and aligning with the encoder's feature space. This effectively resolves the issue of fusion training in vision Transformer~\cite{dosovitskiy2020image} (ViT) architectures becoming trapped in local optima.
\end{itemize}

The remainder of this paper is organized as follows. Section~\ref{sec2} reviews deep learning-based fusion methods, current downstream task-driven fusion approaches, and introduces some basic pretrained network frameworks. Section~\ref{sec3} provides detailed descriptions of the various design details of our work and the relevant theoretical derivations. Section~\ref{sec4} presents both quantitative and qualitative analysis results of MaeFuse on public datasets, along with ablation studies and extended analyses of our method's performance. Finally, Section~\ref{sec5} summarizes our approach and offers a forward-looking perspective.

\section{Related Work}\label{sec2}
In this section, we introduce some typical works related to our method. Firstly, we examine notable deep learning approaches in infrared and visible image fusion. Subsequently, we discuss research focusing on downstream task-driven image fusion. Lastly, we explore various pretrained models utilized for feature representation.

\subsection{Deep Learning-based IVIF}

The development of deep learning has brought many novel methods to image fusion. Infrared and visible image fusion methods based on deep learning can generally be divided into post-fusion frameworks and pre-fusion frameworks. 

For the post-fusion framework, models are often pre-trained using large-scale datasets, so that the encoder obtains good image feature information. Li \emph{et al. } firstly introduced the pretrained fusion model known as DenseFuse~\cite{DenseFuse}, which is comprised of three parts: an encoder layer, a fusion layer, and a decoder layer. In the fusion layer, they applied strategies such as element-wise addition or $\ell_1$-norm, and implemented dense connections within the encoder layer to achieve effective fusion outcomes. Later, they also developed a multi-scale architecture and nested connections to enhance the extraction of more detailed features~\cite{NestFuse}~\cite{RFN-Nest}. Zhao \emph{et al. } created a novel encoder designed for multi-scale decomposition~\cite{DIDFuse}, aimed at extracting both detailed and background features. In a similar vein, Tang \emph{et al. } applied Retinex theory to improve nighttime image fusion results by employing several encoders~\cite{divfusion}, which separate the illumination and reflection components of visible images. However, the methods mentioned above adopted manually designed fusion strategies, which limit their flexibility and fusion performance. Subsequently, Xu \emph{et al. } used a classification saliency-based rule for image fusion~\cite{xu2021classification}, allowing the fusion strategy to be learned as well. Recently, some teams have explored fusion using more information scales based on this structure~\cite{li2024conti}, avoiding the shortcomings of only dividing information into low-frequency and high-frequency. Additionally, other teams have incorporated textual information control at the feature level~\cite{cheng2025textfusion}, making the fusion process more controllable.

Pre-fusion frameworks eliminate the need for manually designed loss functions, network architectures, and learning paradigms. Since fused images lack ground truth, researchers have developed various loss functions based on the characteristics of image fusion to represent the most basic fusion information, thereby providing targeted guidance for training. Based on common intensity and gradient losses, Ma \emph{et al. } designed a fusion loss based on a significant target mask, used to selectively fuse target and background areas~\cite{STDFusionNet}.

Moreover, considering variations in illumination, they designed an illumination-aware loss function~\cite{Tang2022PIAFusion}. Structural similarity loss~\cite{rxdnfuse} and perceptual loss~\cite{ma2020infrared} were also introduced to restrict fusion outcomes and avoid distortion of structural information while ensuring good visual perception. Beyond loss functions, numerous innovative network architectures have been developed, such as residual blocks~\cite{STDFusionNet}, aggregated residual dense blocks~\cite{rxdnfuse}, and gradient residual dense blocks~\cite{SeAfusion}, along with fusion modules including cross-modality differential aware fusion modules~\cite{Tang2022PIAFusion}, interaction fusion modules~\cite{unsupervised}, global spatial attention modules~\cite{SuperFusion}, and biphasic recurrent fusion modules~\cite{ReCoNet}, to ensure the comprehensiveness of information in the fusion results. Researchers have also introduced some new learning paradigms, such as generative adversarial mechanisms and unsupervised learning. Ma \emph{et al. } were the first to introduce generative adversarial networks into the field of image fusion with the development of FusionGAN~\cite{FusionGAN}. This model uses a discriminator to force the generator to retain more texture details from visible images, although it does not adequately consider information from infrared images. The successors to FusionGAN, such as DDcGAN~\cite{DDcGAN}, AttentionFGAN~\cite{AttentionFGAN}, and SDDGAN~\cite{zhou2021semantic}, have designed dual discriminators to address this issue, avoiding the modal imbalance caused by a single discriminator. Unsupervised learning methods~\cite{jung2020unsupervised,cheng2023mufusion} are also employed. Some teams have proposed FusionBooster~\cite{cheng2024fusionbooster} to compensate for the loss of detailed content in fused images, enabling the fusion results to recover relevant texture information.

In recent years, due to the Transformer's superior long-range learning capability, the Vision Transformer (ViT) has gradually replaced CNN as the basic framework for vision. As a result, some Transformer-based fusion models, such as IFT~\cite{ift}, AFT~\cite{aft}, SwinFuse~\cite{wang2022swinfuse},and SwinFusion~\cite{SwinFusion}, have been developed to fully explore the long-range dependencies in source images. Considering that infrared and visible light images often exhibit various degrees of misalignment in practical applications, the latest methods (such as RFNet~\cite{xu2022rfnet}, UMF-CMGR~\cite{unsupervised}, ReCoNet~\cite{ReCoNet}, and SuperFusion~\cite{SuperFusion}) incorporate an alignment module before the fusion module to correct misalignments in the source images. Additionally, some methods, including PMGI~\cite{zhang2020rethinking}, IFCNN~\cite{zhang2020ifcnn}, U2Fusion~\cite{U2Fusion}and DeFusion~\cite{Defusion}, handle various image fusion tasks in a unified manner, as there are commonalities among these tasks. Particularly, U2Fusion~\cite{U2Fusion} trains a unified model to handle multiple fusion tasks, promoting cross-fertilization among different fusion tasks. All the methods above mainly focus on integrating the complementary information of the two modalities and enhancing the information in the fused images. A good visual result is their common pursuit. However, they generally cannot preserve the high-level visual information of the fused images to help us highlight the interested objects.

\subsection{Downstream Task-Driven IVIF}

To enhance the acquisition of high-level visual information for improved image fusion, Tang \emph{et al. } initially employed semantic segmentation tasks as drivers for image fusion~\cite{SeAfusion}. In a similar vein, Liu \emph{et al. } utilized object detection for the same purpose~\cite{TAL}. However, the limitation of labeled fusion datasets hampers the ability to provide semantic feedback to the fusion network through backpropagation. Addressing this, the work of~\cite{PSfusion} introduces a technique to infuse high-level semantic information at the feature level within the fusion network. This is executed by co-training the encoder with both the downstream and fusion tasks. On the other hand, Zhao \emph{et al. } recommended the adoption of meta-learning techniques~\cite{Metafusion}, allowing the fusion network to progressively assimilate both low-level and high-level visual information. Meanwhile, to better utilize image segmentation information, Liu \emph{et al. } injected features from the segmentation task back into the fusion network~\cite{liu2023multi}, guiding the image to more effectively integrate segmentation information. Recently, Zhang \emph{et al. } combined object detection with diffusion models~\cite{zhang2024e2e}, aiming to achieve better fused images by leveraging the object detection's ability to capture information and the powerful image generation capabilities of diffusion models. These methods have brought fresh air to the field of image fusion, and have recently become one of the research hotspots. Nonetheless, these methods have led to increased complexity in both the fusion network and its training strategy, thereby complicating the design and implementation of fusion methods.

\subsection{Pretrained Models for Feature Representation}

A large pre-trained feature model achieves strong generalization and performance for downstream tasks, benefiting low-sample scenarios. For pretrained models, there are mainly two types of architectures: based on convolution and based on Transformers. Within the realm of traditional convolution, VGG~\cite{VGG} initially served as the foundational backbone for encoder architectures. Subsequently, the advent of ResNet~\cite{resnet} addressed the issue of gradient vanishing, enabling deeper network structures for enhanced feature extraction. However, it's important to note that both VGG and ResNet depend heavily on labeled data. To improve feature extraction capability and reduce dependence on labels, SimCLR~\cite{simCLR} uses contrastive learning to help the model obtain useful image representations. Regarding ViT, two primary architectures stand out: SimMIM~\cite{SimMIM} and MAE~\cite{Mae}. Both employ a masking strategy for self-supervised training during their pretraining phase. Notably, the MAE pretrained encoder excels in feature extraction, effectively bridging low-level and high-level visual information. Consequently, in our MaeFuse framework, we leverage the MAE pretrained encoder to enrich and enhance the comprehensiveness of feature information in our fusion results.

\section{Method}\label{sec3}

In this section, we first use mathematical derivations to demonstrate that the encoder requires high-level semantic information. Then, we clarify how to utilize the MAE's pretrained encoder for feature fusion. Finally, we provide a detailed exposition of the guided training strategy, focusing on two aspects: the loss function and the training process.

\begin{figure*}[h] 
  \centering 
  \includegraphics[width=0.98\textwidth]{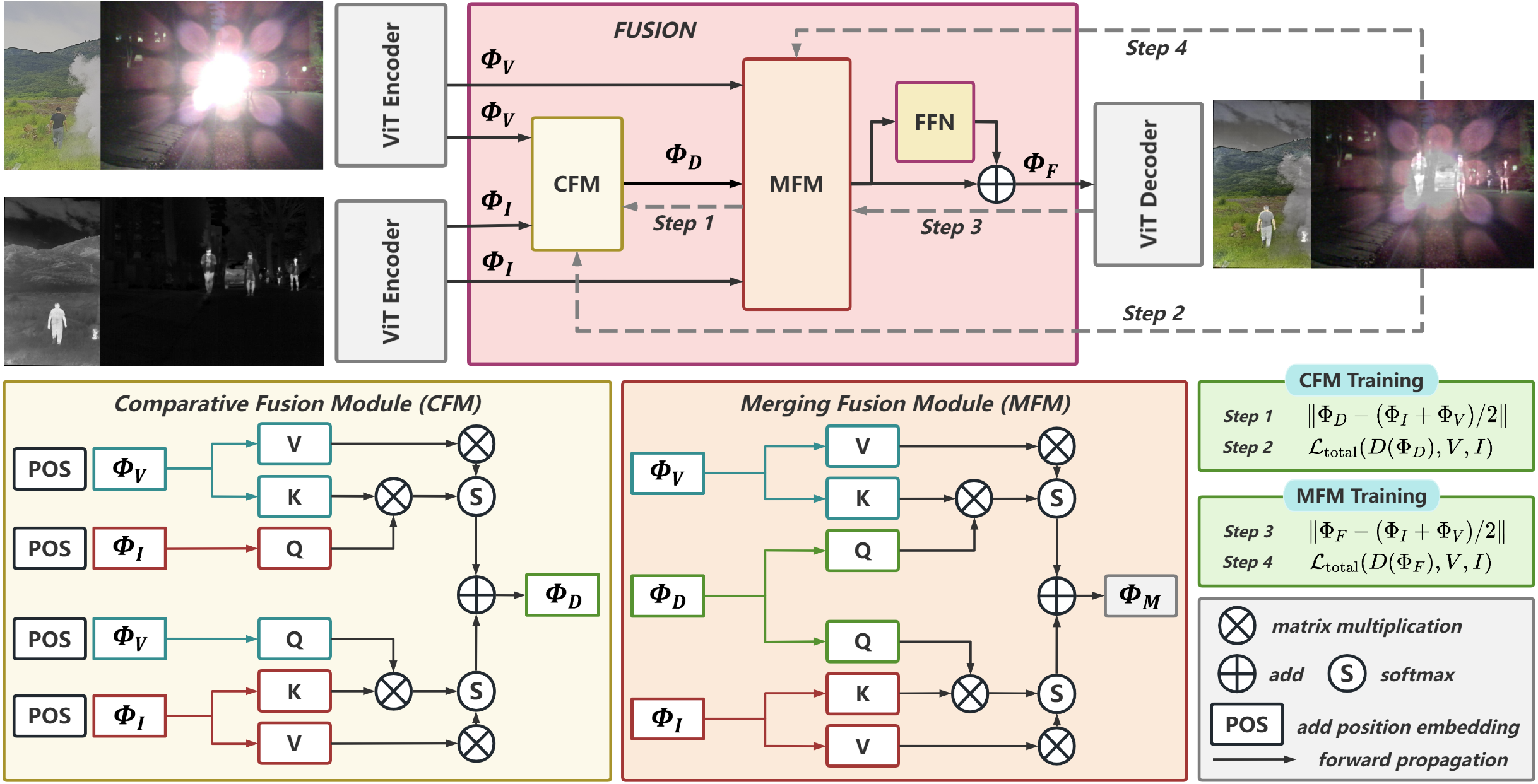} 
  \caption{Workflow of our proposed MaeFuse. The upper part describes the overall architecture of the network, while the lower part elaborates on the content of the CFM and MFM structures. CFM cross-learning retains useful information, and MFM fuses enriched detail information based on the output content of CFM as a reference. The FFN employs merely two layers of fully connected neural networks, aimed specifically at enhancing the model’s capacity for non-linear learning.}
  \vspace{-15pt}
  \label{fig:net}
\end{figure*}

\subsection{Proposition} \label{preliminaries}
First, we aim to use formulas to demonstrate that only when the encoder obtains sufficient semantic information will our final fusion result highlight adequate semantic information. We define the contents of the labels: \(V_f\) represents the features corresponding to the visible image obtained by the encoder, \(I_f\) represents the features corresponding to the infrared image obtained by the encoder, and \(\hat{F}\) represents the fusion result obtained by the existing network. The result fused through the existing network is a definite value, with the probability of 1,
\begin{align}
P(\hat{F}|V_f, I_f) = 1. \label{eq:1}
\end{align}
We aim to establish the correlation between the features obtained from the two modalities through the encoder and the global high-level semantic information. Therefore, we define \(S\) to represent the inherent high-level semantic information of the two modalities. The following formula represents the probability of obtaining the inherent high-level semantic information of the two modalities given the infrared and visible image features,
\begin{align}
P(S|V_f, I_f).
\end{align}
Due to Equation \(\eqref{eq:1}\), we can derive the following:
\begin{align}
P(S|V_f, I_f) &= P(S, \hat{F}|V_f, I_f) + P(S, \neg\hat{F}|V_f, I_f), \notag\\
P(S, \neg\hat{F}|V_f, I_f) &= P(S| \neg\hat{F}, V_f, I_f)(1 - P(\hat{F}|V_f, I_f)), \notag\\
P(S|V_f, I_f) &= P(S, \hat{F}|V_f, I_f),\\
P(S, \hat{F}|V_f, I_f) &= P(S|\hat{F}, V_f, I_f)P(\hat{F}|V_f, I_f), \notag\\
P(S, \hat{F}|V_f, I_f) &= P(S|\hat{F}, V_f, I_f), \notag\\
P(S, \hat{F}|V_f, I_f) &= P(S|\hat{F}).
\end{align}
From the above equations, we can obtain the following:
\begin{align}
P(S|V_f, I_f) &= P(S, \hat{F}|V_f, I_f) = P(S|\hat{F}) \leq P(S|F).\label{eq:1_equation}
\end{align}
Where \(F\) represents the optimal fusion effect, and thus, the optimal fusion effect undoubtedly preserves the most high-level semantic information.

By combining the above equations, we can consider that to achieve the best fusion result, the encoder must effectively capture high-level semantic information from the image features. Furthermore, this information must be efficiently utilized within the fusion structure to attain a satisfactory fusion outcome.

\subsection{Network Architecture} \label{preliminaries}

The proposed MaeFuse primarily consists of two architectures: 1) A pretrained encoder and decoder based on MAE, utilizing the encoder weights from MAE to directly obtain our feature vectors. 2) A learnable two-layer fusion network. We illustrate the proposed method in Fig.~\ref{fig:net}.

\textbf{MAE.} Since MAE is a type of Vision Transformer (ViT), we first briefly review the overview of ViT. For a 2D image \(I\), ViT represents \(I\) as a sequence of 1D patch embeddings \( T = \{ t_i \mid i = 1, \ldots, n \} \), where \(n\) denotes the total number of patches. Each patch embedding \(t_i\) corresponds to a fixed-size section of \(I\). Additionally, positional embeddings are added to these patch embeddings to preserve the positional information of the patches. Besides positional embeddings, there is also a classification patch embedding, \( [ CLS ] \), included as an additional learnable embedding, which serves as the global image representation. The complete set of embeddings is referred to as a set of tokens \( T = \{ t_i \mid i = CLS, 1, \ldots, n \} \).

The tokens \( T \) are then passed through the Transformer encoder, which is composed of multiple Transformer layers stacked together. Each Transformer layer contains Layer Normalization (LN) layers, Multi-Head self-Attention (MHA) modules, and Multi-Layer Perceptron (MLP) blocks. The output of the tokens from the \((l+1)\)-th Transformer layer can be expressed as:
\begin{align}
T^{l+1} &= \text{MHA}(\text{LN}(T^l)) + T^l, \notag \\
T^{l+1} &= \text{MLP}(\text{LN}(T^{l+1})) + T^{l+1}.
\end{align}%
Specifically, within the multi-head self-attention (MHA) block, tokens are transformed into queries, keys, and values:
\begin{align}
Q^{l+1} &= T^l \cdot W_q^{l+1}, \notag\\
K^{l+1} &= T^l \cdot W_k^{l+1}, \notag\\
V^{l+1} &= T^l \cdot W_v^{l+1},
\end{align}%
where \( W_q^{l+1} \), \( W_k^{l+1} \), and \( W_v^{l+1} \) are learnable weights.

MAE is based on this structure and conducts self-supervised training. Specifically, we randomly select some tokens to mask \( M = \{m_i \mid m_i \in \{0, 1\},  i = 1, \ldots, n\} \), and then use the unmasked parts to predict the masked areas. Since we need to recover the entire image from only 25\% of the remaining regions, we must use a small amount of information to acquire semantic information to guide the recovery of the masked areas. This allows our encoder to obtain good high-level semantic information. Additionally, in the image recovery process, we need to match the colors and textures with the remaining parts of the image, so the encoder also acquires valuable low-level visual information. Therefore, good omni features that facilate the low-level and high-level task simulantoursly can be obtained through self-supervised training with random masking.

\textbf{Encoder.}  Here we utilize the MAE (large) architecture, characterized by 24 layers of ViT blocks. To process both infrared and visible images, we employ a unified encoder. The visible images are initially transformed into YCrCb format, with only the Y channel being used for input. This approach enables us to project information from both modalities into a singular, coherent feature space.

A brief notation is introduced for clarity. The input images, namely the visible image and the infrared image, are denoted as \( V \in \mathbb{R}^{H \times W } \) and \( I \in \mathbb{R}^{H \times W } \), respectively. Given that a singular encoder is utilized for both, it is referred to as \(  \mathbb{E}(\cdot) \) in our discussion.

Our encoding process can be described as converting the inputs \(\{V, I\}\) from visible and infrared images into corresponding image features \(\{\Phi_V, \Phi_I\}\)
\begin{align}
\Phi_V = \mathbb{E}(V), \quad \Phi_I = \mathbb{E}(I). 
\label{eq:encoder}
\end{align}%

\textbf{Fusion Layer.} In the fusion process, our objective is for the network to selectively integrate modal information in various areas, drawing on the feature information from both modalities. Essentially, the aim is to ensure that the fusion outcome in each region is enriched with a higher density of information. The first part of the fusion layer is the Comparative Fusion Module (CFM), which primarily facilitates the interactive learning of features from two modalities through two symmetric cross-attention networks. The output of CFM, represented by \(\Phi_D\), predominantly captures essential contour information from both modalities. However, it is important to note that during the cross-learning process, this module often loses a significant amount of detail information.

To mitigate the problem of the Comparative Fusion Module (CFM)'s output lacking fine details, we incorporated the Merging Fusion Module (MFM). Within this framework, the feature vector \(\Phi_D\), produced by the CFM, acts as a guide for contour features, aiding in the re-fusion of the initial encoded features. The original encoded features \(\Phi_V\) and \(\Phi_I\) from the two modalities are then compared with \(\Phi_D\). In regions where the original encoding closely aligns with \(\Phi_D\), the information is retained, while dissimilar regions are de-emphasized. This method ensures that the resulting fusion feature vector not only retains the contour information of \(\Phi_D\) but also enriches it with more detailed information. And the FFN structure has strong non-linear learning capabilities, allowing us to better align our fusion domain with the domain of the decoder.

Mathematically, CFM is denoted as \( \mathbb{C}(\cdot) \), MFM as \( \mathbb{M}(\cdot) \) and FFN as \( \mathbb{FFN}(\cdot) \). \(\Phi_D\) represents the output content of CFM, \(\Phi_M\) represents the output content of MFM, while \(\Phi_F\) is the final output result. Thus, we have
\begin{align}
\Phi_D &= \mathbb{C}(\Phi_I, \Phi_V), \notag \\
\Phi_M &= \mathbb{M}(\Phi_I, \Phi_V, \Phi_D), \notag\\
\Phi_F &= \Phi_M + \mathbb{FFN}(\Phi_M).
\label{eq:fusion}
\end{align}%

\textbf{Decoder.} For the ViT architecture, a simple few-layer decoder can achieve very good reconstruction results, so we choose 4-layer ViT blocks as the decoder. Here, \( \mathbb{D}(\cdot) \) represents the decoder and we have
\begin{align}
F = \mathbb{D}(\Phi_F).
\label{eq:decoder}
\end{align}%

Here, we have removed the mask from MAE, and we directly use the $\mathcal{L}_1$ loss function to guide the decoder in reconstructing the original input image.

\begin{figure}[h] 
  \centering 
  \includegraphics[width=\columnwidth]{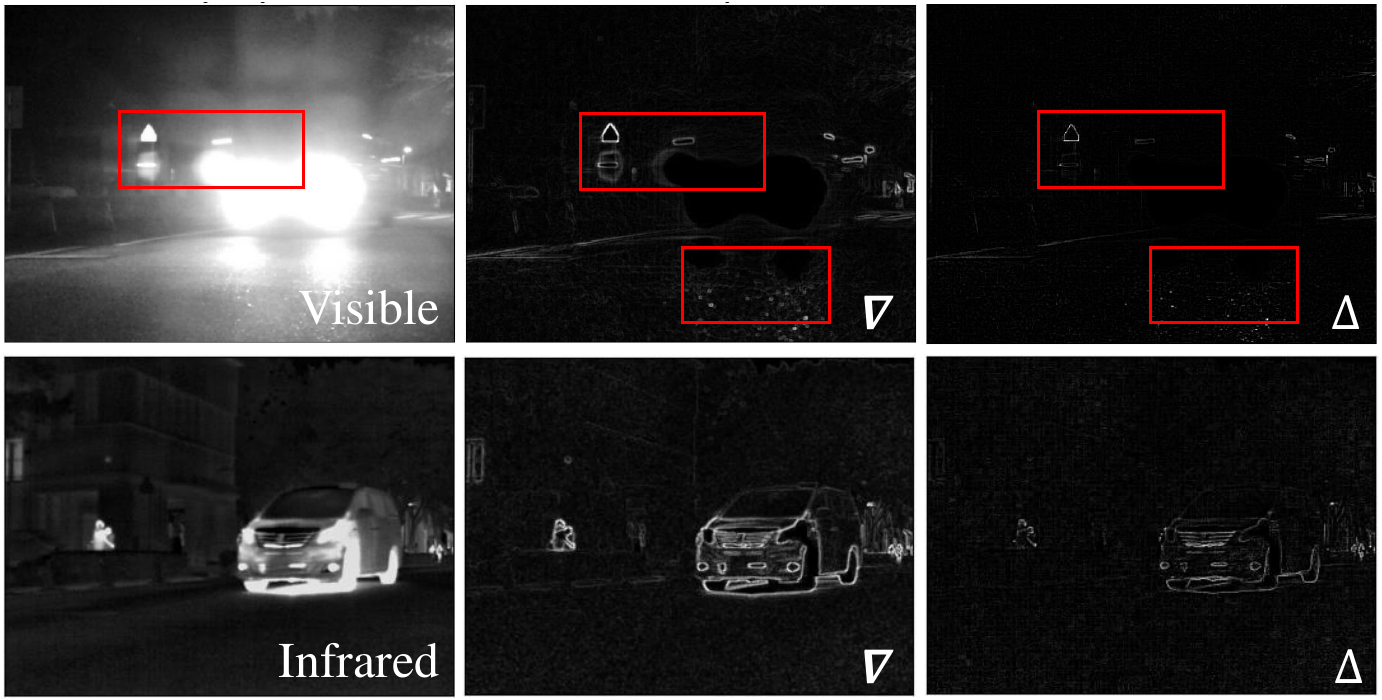} 
  \caption{The image `00909N' scene is from the MSRS dataset. The first row shows visible images, while the second row shows infrared images. The first column contains the original images, the second column contains gradient images, and the third column contains second derivative images. \(\nabla\) is the Sobel operator, and \(\Delta\) is the Laplacian operator.}
  \vspace{-15pt}
  \label{fig:loss_cmp}
\end{figure}

\subsection{Loss Function}
Firstly, we need to train the decoder to acquire the ability to recover the original image from image features:
\begin{align}
\mathcal{L}_{\text{decoder}} &= \left\| \mathbb{D}(\mathbb{E}(I)) - I \right\|.
\end{align}

Here, we lock the encoder \(  \mathbb{E}(\cdot) \) and then apply $\mathcal{L}_1$ loss between the decoder's \(  \mathbb{D}(\cdot) \) output image and the original image to update the decoder's weight information.

In our previous formula derivation~\ref{eq:1_equation}, we know that if our fusion results benefit downstream tasks, then our encoder needs to acquire sufficient semantic information, and our fusion layer also needs to reasonably learn and utilize this information. For the encoder, we addressed the issue by utilizing the pretrained MAE encoder. As for the fusion layer, we use texture loss to guide the network to retain semantic information. The gradient information obtained from the first-order derivative of the image also contains semantic texture information to some extent, so our loss function uses the $\mathcal{L}_1$ loss function, the gradient loss function along with the laplacian loss:
\begin{align}
\mathcal{L}_{\text{total}} &= \mathcal{L}_{\text{int}} + \alpha \mathcal{L}_{\text{grad}} + \beta \mathcal{L}_{\text{laplacian}} \notag, \\
\mathcal{L}_{\text{int}} &= \frac{1}{HW} \left\| F - \max(V, I) \right\| \notag, \\
\mathcal{L}_{\text{grad}} &= \frac{1}{HW} \left\| \left| \nabla F \right| - \max\left( \left| \nabla V \right|, \left| \nabla I \right| \right) \right\| \notag,\\
\mathcal{L}_{\text{laplacian}} &= \frac{1}{HW} \left\| \left| \Delta F \right| - \max\left( \left| \Delta V \right|, \left| \Delta I \right| \right) \right\|.
\label{eq:loss_total}
\end{align}%

Generally, fusion work commonly uses gradients to represent texture information, as proposed by the GTF~\cite{GTF} method, which has become a standard loss function in fusion. However, these texture information might include unwanted elements (such as overexposure, smoke). Here, we explored the effect of second derivatives on fusion. Since object information generally has distinct boundaries, the gradients can change dramatically; whereas for overexposure and similar effects, there is a gradual change, and the magnitude of gradient changes are not as severe. After calculating the second derivatives, these values will be very small, allowing us to reduce the interference from non-object contour information to some extent. As shown in Fig.~\ref{fig:loss_cmp}, combining the use of gradients and second derivatives can achieve a better fusion effect.

\begin{figure*}[t] 
  \centering 
  \includegraphics[width=0.98\textwidth]{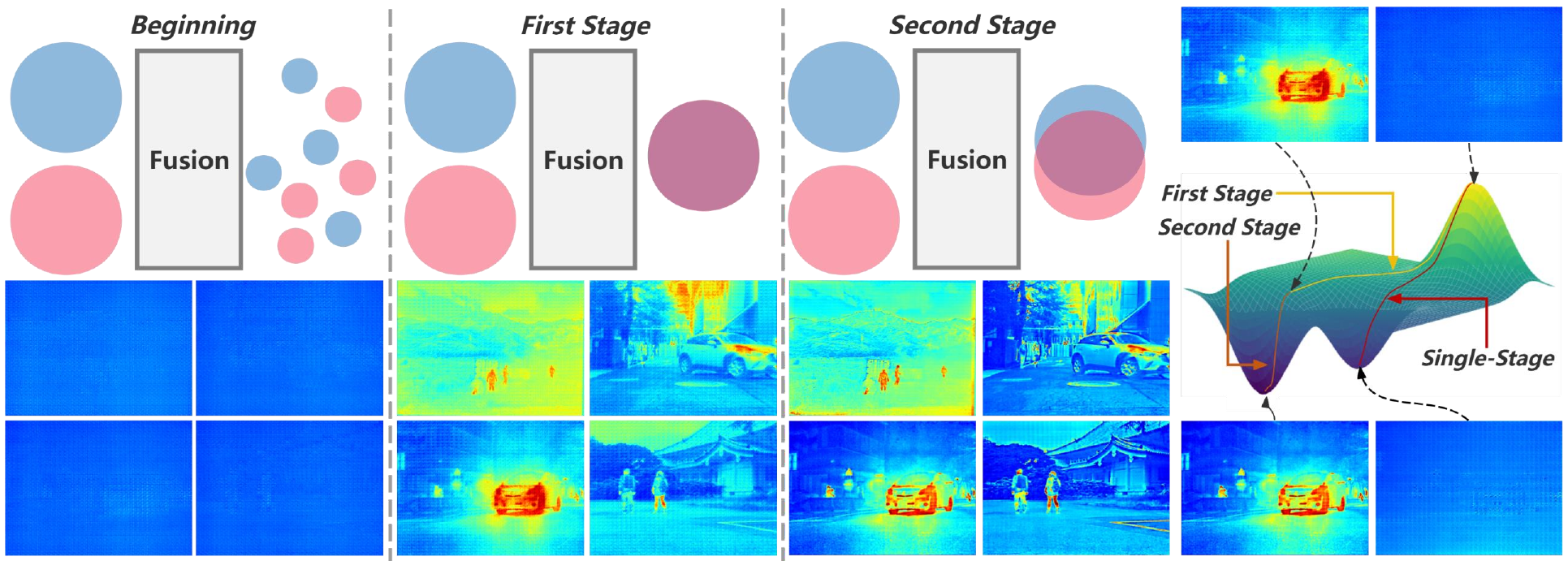} 
  \caption{The schematic diagram illustrates our two-stage training approach. The left of each fusion module displays the features obtained by the encoder, whereas the right shows the features obtained by the fusion layer. The first stage involves aligning the feature domains of the fusion layer and the encoder. The second stage progresses with training using a fusion loss function Eq. (\ref{eq:loss_total}). This two-stage training strategy is designed to effectively circumvent the issue of becoming trapped in local optima.}
  \vspace{-15pt}
  \label{fig:train3}
\end{figure*}

Certainly, these loss functions are designed to globally optimize the image. Since we have already included high-level semantic information, we do not need related loss functions. Before calculating the fusion loss, we first need to align the output feature domain of the fusion layer with the feature domain of the encoder. Here, we use the mean of the two modalities as the baseline for training
\begin{align}
\mathcal{L}_{\text{CFM-align}} = \left( \frac{\left( \Phi_I + \Phi_V \right)}{2} - \Phi_D \right)^2  \notag,\\
\mathcal{L}_{\text{MFM-align}} = \left( \frac{\left( \Phi_I + \Phi_V \right)}{2} - \Phi_F \right)^2 .
\label{eq:loss_align}
\end{align}

By using the alignment function mentioned above, we can ensure that the current results are comparable to the mean fusion, thus aligning the fusion domain with the encoder's feature space. On this basis, we can then perform fusion training to achieve a good result.

\subsection{Guided Training Strategy}

Guided learning is an approach that focuses on achieving specific objectives to enhance learning efficiency. In our fusion task, we faced two primary challenges: limited data availability and the emergence of block effects with discordant image information when directly fusing features from the MAE encoder. To tackle these issues, we implemented guided training. This method sets predetermined objectives to steer the training process. Employing this strategy allows us to effectively address both challenges. It ensures the model avoids settling at local optima and achieves more accurate results.

Essentially, our guided training strategy is a variant of knowledge distillation, where we use existing knowledge to quickly teach an uninitialized model structure to learn on a relatively good prior basis. For our fusion work, this is akin to using the encoder's output to guide the initialization of the fusion layer's weights. The reason we use the mean fusion of encoder output features as a guide here draws from the idea of residual connections; our initialization is similar to a direct residual connection. As we gradually train the fusion layer, we can ensure that the minimum performance is no worse than the result of mean fusion. Another point is the desire for the fusion layer's output domain to align with the decoder's domain, making the fusion training more effective. Broadly speaking, for infrared and visible image fusion, or other tasks without ground truth (GT), we often use an approximate loss function for training. The convergence ability of this type of loss is not as good as that with GT, and it inherently carries some bias. By using guided learning, we can effectively inject certain priors into the model, thereby reducing the final deviation from the actual GT, making it closer to the real GT. Specific steps are detailed in the following two subsections.

\subsubsection{Two-Stage Training}
The MAE encoder can represent image data in a feature domain, where we need to fuse features of two modalities. Therefore, we first need to align the feature domain output by the fusion layer with the feature domain of the encoder. This alignment is our goal for guided learning. Once our feature domains are aligned, we can use a fusion loss function to promote the integration of information from the two modalities. Ignoring this phased training approach may lead to getting trapped in local optima. Our training strategy schematic is shown in Fig.~\ref{fig:train3}.

We first define our guiding objective, which is the mean of the feature vectors of the two modalities. We calculate the least squares of the fusion network's output features with this target. This process helps to quickly align the feature domain output by the fusion layer with the feature domain of the encoder. Afterwards, we use an image texture loss function to guide the fusion effect towards a direction with richer texture. See Algorithm~\ref{alg:Two-Stage} for the pseudocode.

\begin{algorithm}[t]
    \caption{The Proposed Guided Training Strategy}
    \label{alg:Two-Stage}
    \textbf{Input}: \(\Phi_V\) and \(\Phi_I\)\\
    \textbf{Output}: \(\Phi_F\)
    \begin{algorithmic}[1] 
        \STATE Let $t=0$.
        \STATE Initialize \(\Phi_F\).
        \WHILE{$t < 100$}
            \IF {$t < 20$}
                \STATE Compute loss with Eq. (\ref{eq:loss_align}) between \(\Phi_F\) and \((\Phi_V + \Phi_I) / 2\).
            \ELSE
                \STATE Compute loss of \(D(\Phi_F)\) with Eq. (\ref{eq:loss_total}). 
            \ENDIF
            \STATE Update MFM's weight based on the computed loss.
            \STATE $t = t + 1$.
        \ENDWHILE
        \STATE \textbf{return} \(\Phi_F\)
    \end{algorithmic}
\end{algorithm}

\begin{figure}[h] 
  \centering 
  \includegraphics[width=\columnwidth]{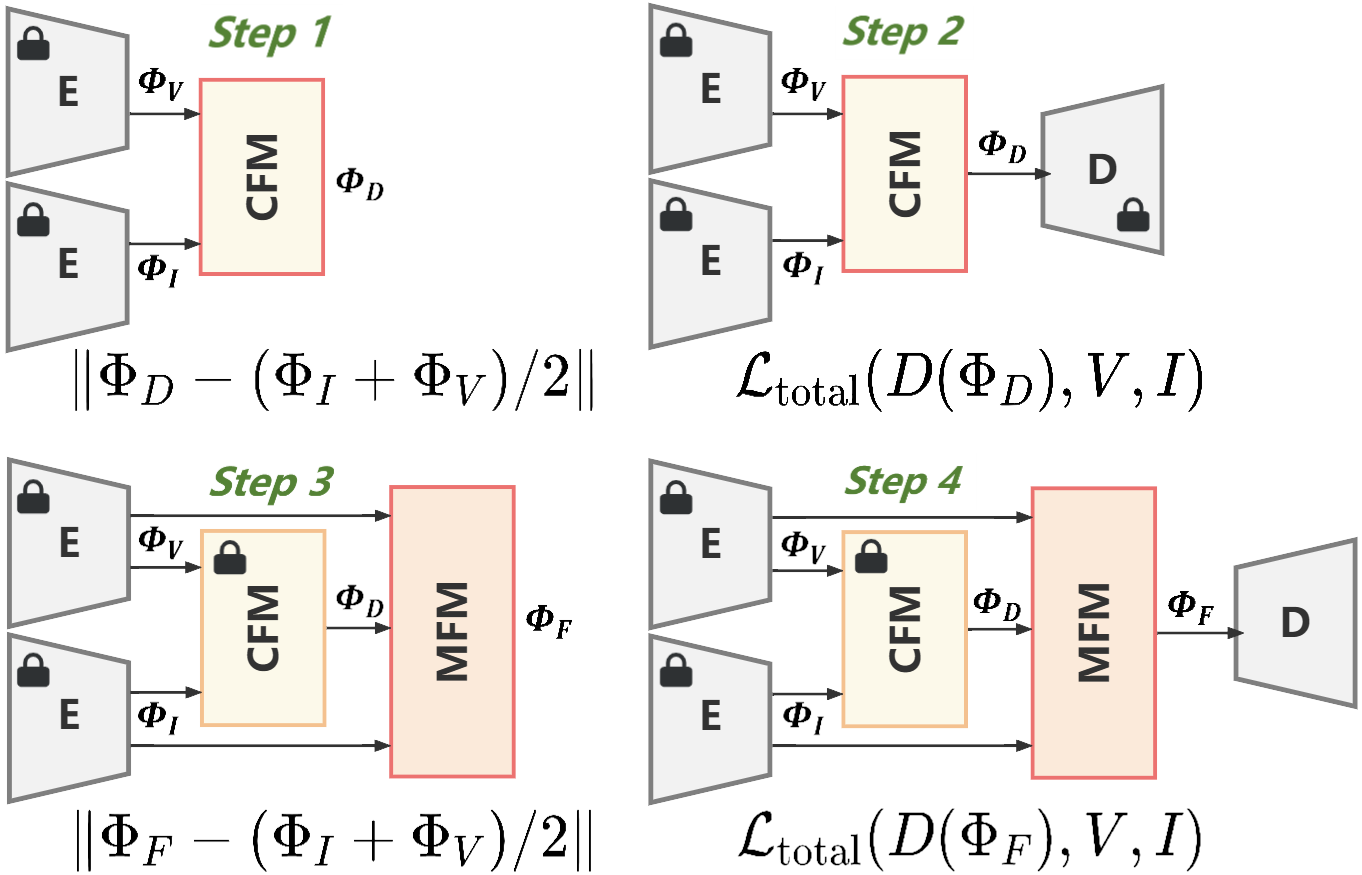} 
  \caption{Illustration of the proposed guided training process. The first and second rows represent the training process of CFM and MFM, respectively. When conducting training for each module, there are two steps: modality alignment training based on the average of encoder and training based on the fusion loss. These two steps are known as two-stage training.}
  \vspace{-15pt}
  \label{fig:guide}
\end{figure}

\subsubsection{Hierarchical Training of Network Structures}
For the CFM and MFM, we adopt a sequential guided training approach.
Initially, we focus on the CFM, using a two-stage training strategy. This involves directly connecting the CFM's output to the decoder while keeping the MFM inactive. This step prevents the direct connection of the MFM with the encoder from diminishing the cross-learning effect in CFM. Once CFM training is complete, we freeze its weights. Subsequently, we apply a similar two-stage strategy for the MFM, where its input is the output of the now-trained CFM, and its output is linked to the decoder. This approach ensures that the MFM effectively learns the fusion features from the CFM, thereby maximizing the retention of information acquired by the CFM.

\section{Experiments}\label{sec4}

\subsection{Implementation Details}
We conducted experiments on four representative datasets: TNO~\cite{Toet2012ProgressIC}, M\textsuperscript{3}FD~\cite{TAL}, RoadScene~\cite{Xu2020FusionDNAU}, and MSRS~\cite{Tang2022PIAFusion}. The AdamW optimizer was utilized to update the parameters of various model modules. For computations, we employed PyTorch's Automatic Mixed Precision (AMP). The network was trained using a fixed learning rate of \(1e^{-4}\). Our hyperparameters were set to $\alpha = 1$ and $\beta = 2$, and the number of training epochs corresponded to those specified in Algorithm~\ref{alg:Two-Stage}. The model was trained on the training set of the MSRS dataset and evaluated on the test sets of MSRS, M\textsuperscript{3}FD, TNO, and RoadScene. All experiments were implemented using the PyTorch framework.

\subsection{Evaluation in Multi-modality Image Fusion}
We conducted qualitative and quantitative analyses with nine
state-of-the-art competitors, including CUFD~\cite{CUFD}, DIVFusion~\cite{divfusion}, GAN-FM~\cite{GAN-FM}, 
STDFusionNet~\cite{STDFusionNet}, NestFuse~\cite{NestFuse}, SeAFusion~\cite{SeAfusion}, TarDAL~\cite{TAL}, MetaFusion~\cite{Metafusion}, and PSFusion~\cite{PSfusion}. Among them, methods CUFD, DIVFusion, GAN-FM, STDFusionNet, and NestFuse do not employ downstream tasks for driving. Methods SeAFusion, TarDAL, MetaFusion and PSFusion are semantically driven by downstream tasks, where SeAFusion and TarDAL utilize image detection, and MetaFusion and PSFusion employ image segmentation. Our method does not use downstream tasks for driving.

\begin{figure}[h] 
  \centering 
  \includegraphics[width=\columnwidth]{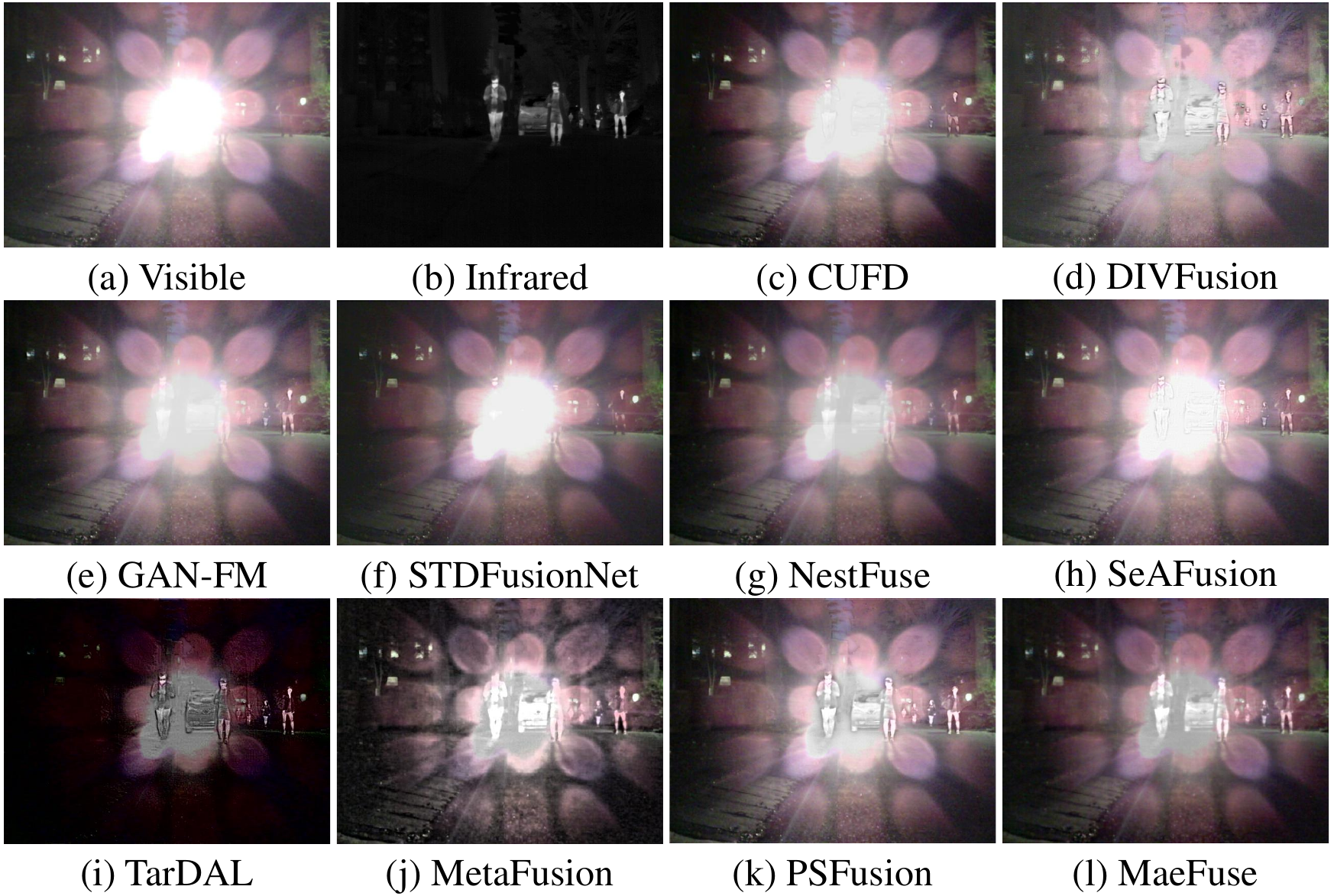} 
  \caption{Qualitative comparison on the `00037N' scene from the MSRS dataset.}
  \vspace{-15pt}
  \label{fig:ex-msrs}
\end{figure}

\begin{figure}[h] 
  \centering 
  \includegraphics[width=\columnwidth]{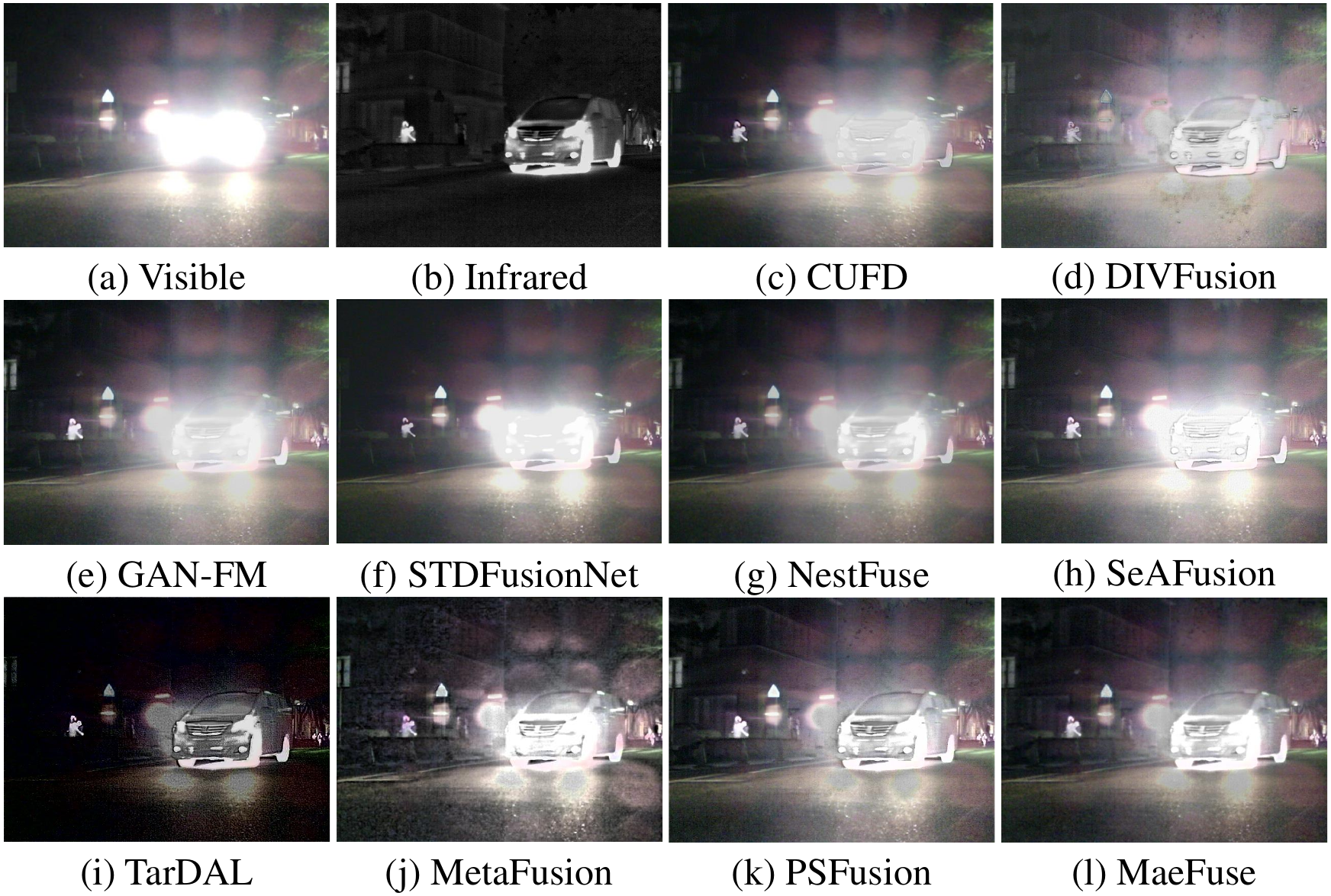} 
  \caption{Qualitative comparison on the `00909N' scene from the MSRS dataset.}
  \vspace{-5pt}
  \label{fig:ex-msrs2}
\end{figure}

\begin{figure}[h] 
  \centering 
  \includegraphics[width=\columnwidth]{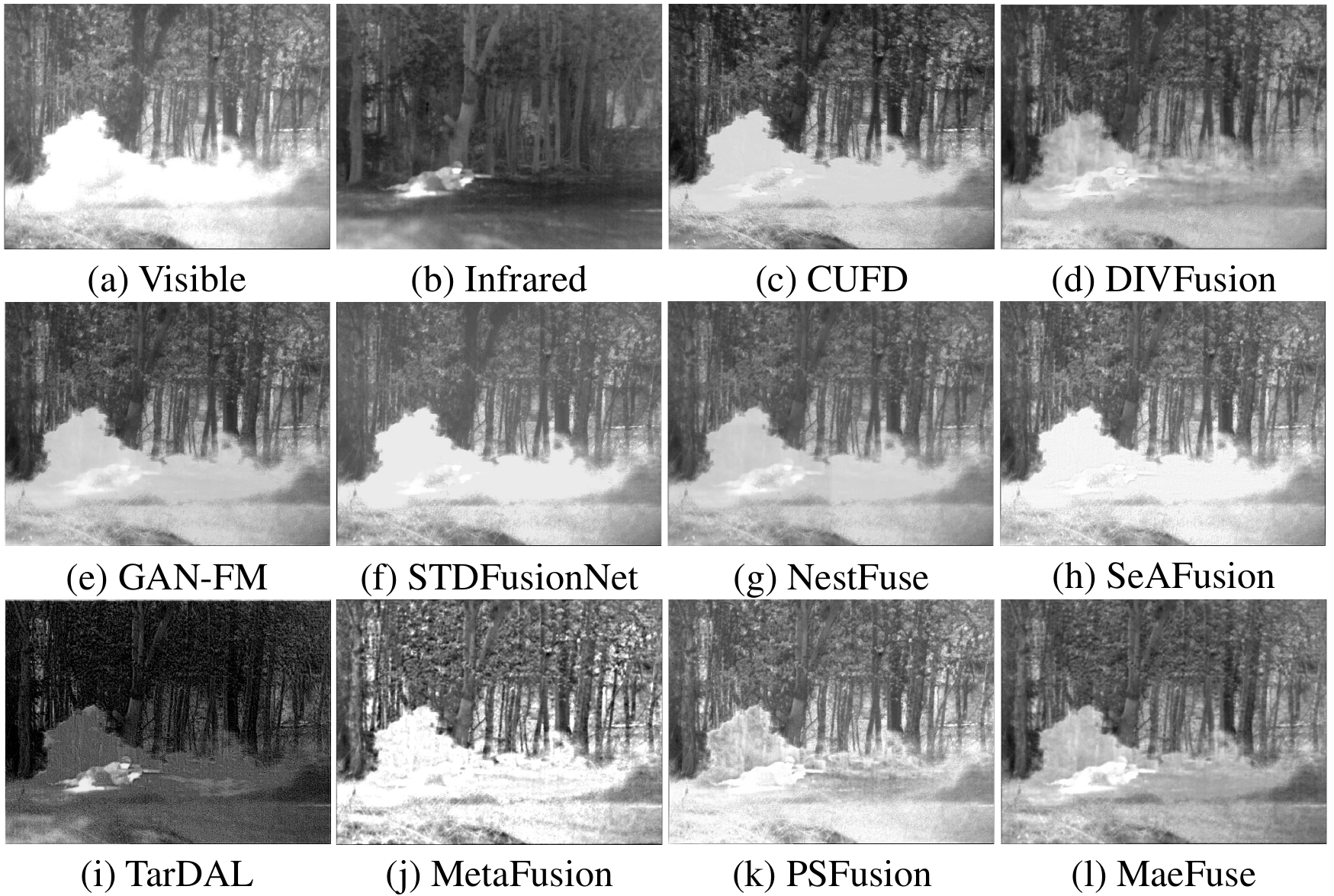} 
  \caption{Qualitative comparison on the `soldier\_behind\_smoke\_1' scene from the TNO dataset.}
 \vspace{-15pt}
  \label{fig:ex-tno}
\end{figure}

\textbf{Qualitative comparisons.} To demonstrate that the MAE pretrained encoder can effectively extract high-level visual information, we categorize the comparative methods into two groups: those without fusion training driven by downstream tasks (c,d,e,f,g) and those with fusion training driven by downstream tasks (h,i,j,k). In the exposed images of the MSRS dataset, we find that methods not driven by downstream tasks fail to effectively represent object information and are easily influenced by high exposure. However, methods driven by downstream tasks can to some extent reveal object information. Our method, not driven by downstream tasks, also shows equally good or even better results, indicating that our encoder can extract and fuse high-level visual information. This is exemplified in Fig.~\ref{fig:ex-msrs} and Fig.~\ref{fig:ex-msrs2}. Similarly, when testing on the TNO dataset, we find that MaeFuse performs better than other works even in conditions with fog obstruction. It is able to maximize the visibility of soldiers behind the fog while retaining the surrounding information, as shown in Fig.~\ref{fig:ex-tno}.

\begin{figure}[h] 
  \centering 
  \includegraphics[width=\columnwidth]{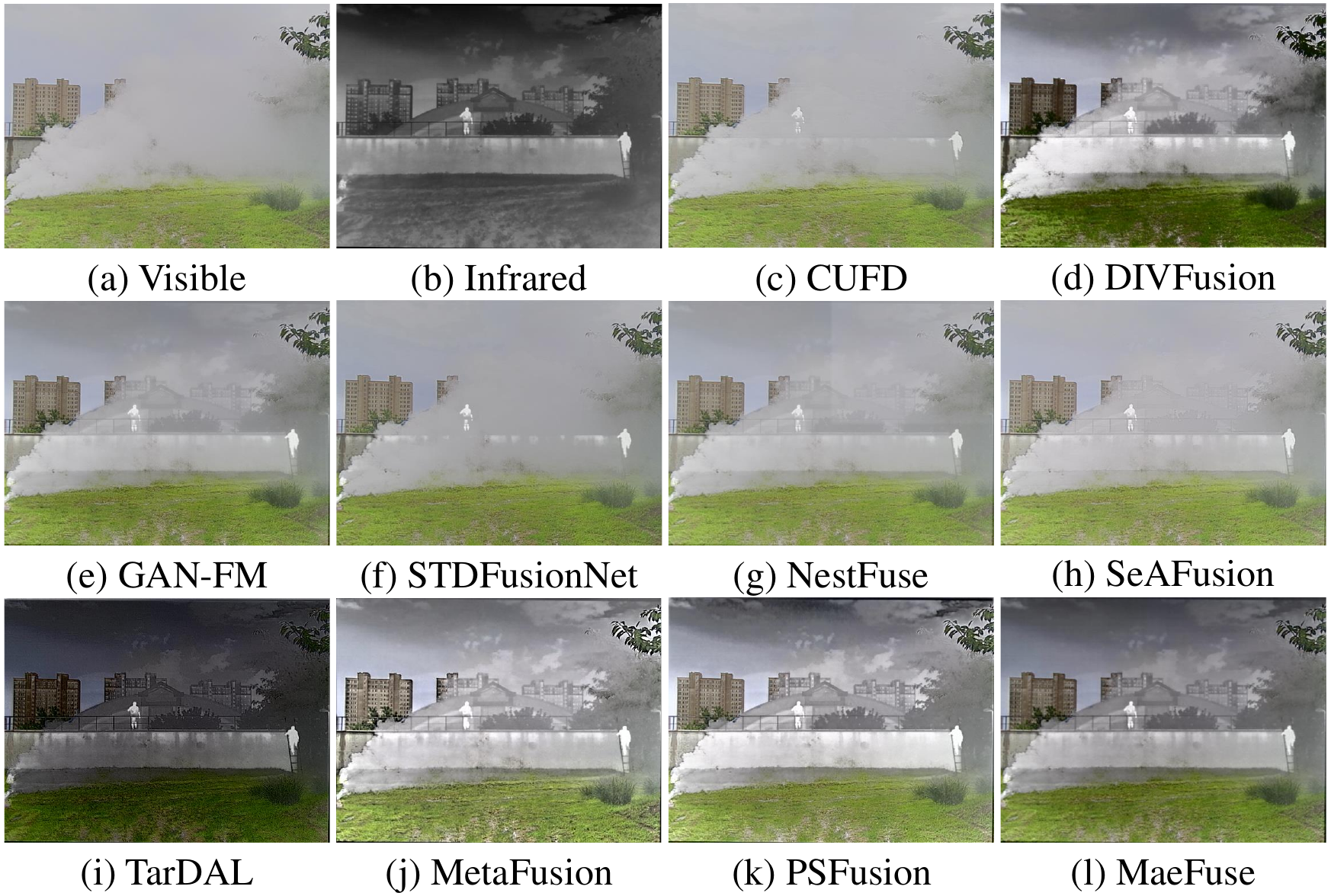} 
  \caption{Qualitative comparison on the `00922' scene from the M\textsuperscript{3}FD dataset.}
  \label{fig:ex-m3fd}
  \vspace{-15pt}
\end{figure}

\begin{figure}[h] 
  \centering 
  \includegraphics[width=\columnwidth]{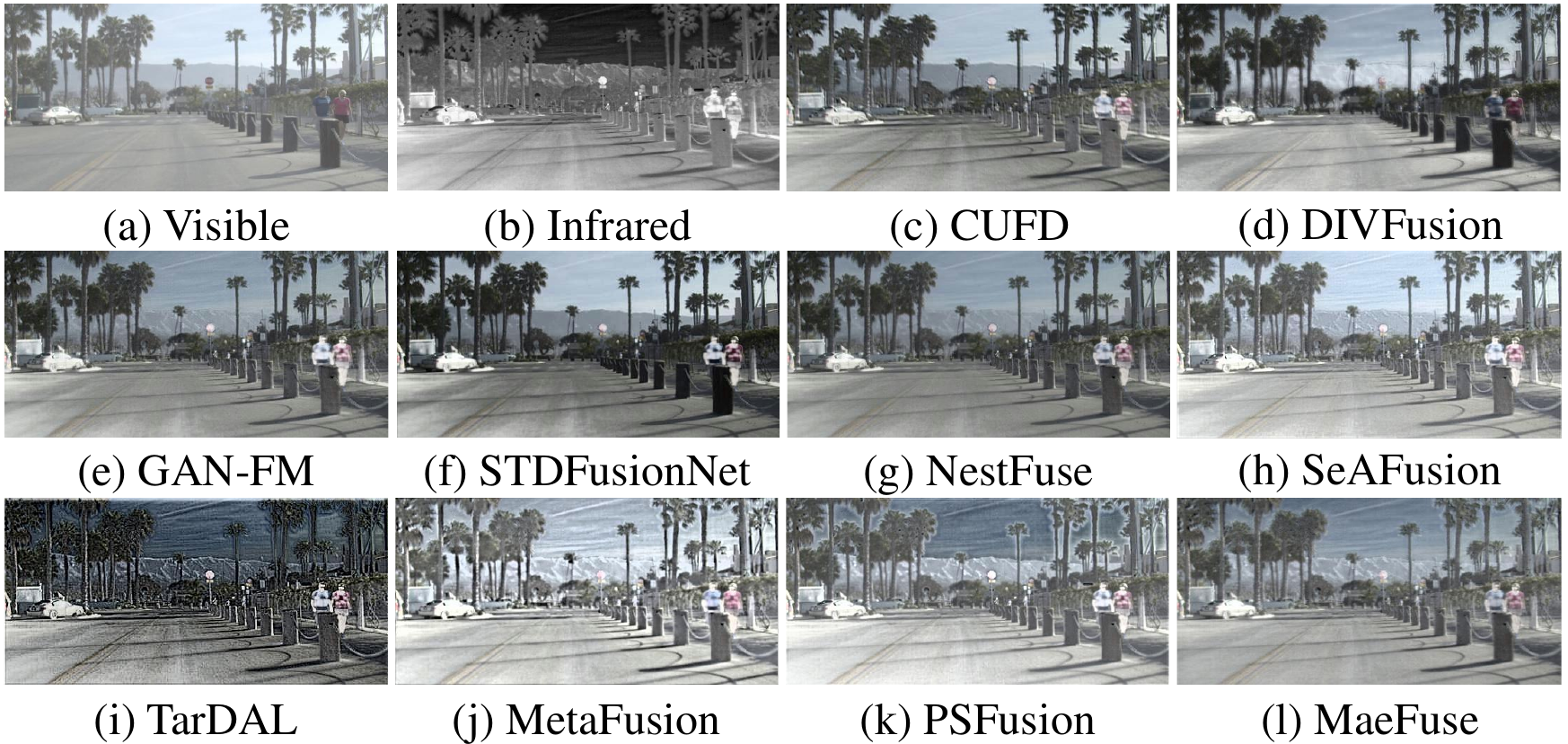} 
  \caption{Qualitative comparison on the `FLIR\_04269' scene from the RoadScene dataset.}
  \label{fig:ex-road}
  \vspace{-5pt}
\end{figure}

Since our encoder establishes a unity between low-level and high-level visual information, the images fused by MaeFuse also maintain texture details well. In the data from M\textsuperscript{3}FD and RoadScene, our fused images fully preserve the texture and color information of the images. Furthermore, our fusion images are not affected by infrared images, thereby avoiding artifacts. Our pictures do not display obvious fusion artifacts, and the final results tend towards a natural image effect. The content is shown in Fig.~\ref{fig:ex-m3fd} and Fig.~\ref{fig:ex-road}.

Overall, for works (c,d,e,f,g) that do not use downstream tasks, there is not a good unification of high-level semantic information and texture information. Works (h,i,j,k) driven by downstream tasks can highlight semantic information under extreme conditions to a certain extent. Among these, we observed that works (h,j,k) driven by segmentation tasks outperform those (i) driven by image detection tasks. Our work, MaeFuse, can easily solve the above issues and achieves results that are on par with or even surpass those driven by segmentation tasks. It is worth mentioning that the PSFusion work uses a custom mask to assist image fusion, which separates object information in the infrared images with the mask, thereby injecting artificial priors into the results. Our work, however, does not use any custom masks for driving, relying solely on the model's inherent pre-trained weight priors for fusion, which demonstrates our model's strong learning capability and potential for future improvement!

\begin{table}[h]
\centering
\caption{Quantitative analysis results on RoadScene~\cite{Xu2020FusionDNAU}.}
\begin{tabular}{@{}l|ccccc@{}}
\toprule
Method  & CC $ \uparrow $ & SCD $ \uparrow $ & PSNR $ \uparrow $ & N\textsuperscript{AB/F} $ \downarrow $ & NLPD $ \downarrow $ \\ \midrule
CUFD~\cite{CUFD}         & 0.582                   & 1.420 & 62.23 & 0.051 & 0.6797 \\
DIVFusion~\cite{divfusion}    & 0.606                   & 1.524 & 61.78 & 0.055 & 0.6793 \\
GAN-FM~\cite{GAN-FM}       & 0.621 & \textcolor{blue}{1.703} & 62.03 & 0.046 & 0.6581 \\
NestFuse~\cite{NestFuse}     & \textcolor{blue}{0.635} & 1.701 & 61.99 & \textcolor{blue}{0.031} & \textcolor{blue}{0.6372} \\
STDFusionNet~\cite{STDFusionNet} & 0.590                   & 1.447 & 60.01 & 0.085 & 0.7379 \\
SeAFusion~\cite{SeAfusion}    & 0.614                  & 1.563 & 61.74 & 0.063 & 0.6604 \\
TarDAL~\cite{TAL}       & 0.528                   & 1.261 & 59.57 & 0.352 & 0.8451 \\
MetaFusion~\cite{Metafusion}   & 0.611                   & 1.547 & 61.81 & 0.152 & 0.7348 \\
PSFusion~\cite{PSfusion}     & 0.628 & 1.696 & \textcolor{blue}{63.10} & 0.065 & 0.6842 \\
MaeFuse      & \textcolor{red}{0.653}  & \textcolor{red}{1.714} & \textcolor{red}{63.92} & \textcolor{red}{0.028} & \textcolor{red}{0.5990} \\ \bottomrule
\end{tabular}
\label{tab: road}
\end{table}

\begin{table}[h]
\centering
\caption{Quantitative analysis results on TNO~\cite{Toet2012ProgressIC}.}
\begin{tabular}{@{}l|ccccc@{}}
\toprule
Method  & CC $ \uparrow $ & SCD $ \uparrow $ & PSNR $ \uparrow $ & N\textsuperscript{AB/F} $ \downarrow $ & NLPD $ \downarrow $ \\ \midrule
CUFD~\cite{CUFD}        & 0.455 & 1.528 & 61.14                   & 0.072 & 0.6160 \\
DIVFusion~\cite{divfusion}    & 0.445 & 1.494 & 59.98                   & 0.148 & 0.6799 \\
GAN-FM~\cite{GAN-FM}       & 0.469 & 1.666 & 61.11                   & 0.073 & 0.5941 \\
NestFuse~\cite{NestFuse}     & 0.488 & 1.726 & \textcolor{blue}{62.02} & \textcolor{blue}{0.032} & \textcolor{blue}{0.5381} \\
STDFusionNet~\cite{STDFusionNet} & 0.428 & 1.440 & 61.46                   & 0.076 & 0.5958 \\
SeAFusion~\cite{SeAfusion}    & 0.482 & 1.728 & 61.39                   & 0.081 & 0.5848 \\
TarDAL~\cite{TAL}       & 0.403 & 1.343 & 59.72                   & 0.392 & 0.7949 \\
MetaFusion~\cite{Metafusion}   & 0.488 & 1.768 & 61.78                   & 0.208 & 0.6669 \\
PSFusion~\cite{PSfusion}     & \textcolor{blue}{0.498} & \textcolor{blue}{1.778} & 61.72 & 0.092 & 0.6316 \\
MaeFuse      & \textcolor{red}{0.524} & \textcolor{red}{1.820} & \textcolor{red}{63.48} & \textcolor{red}{0.030} & \textcolor{red}{0.5272} \\ \bottomrule
\end{tabular}
\label{tab: tno}
\end{table}

\begin{table}[h]
\centering
\caption{Quantitative analysis results on M\textsuperscript{3}FD~\cite{TAL}.}
\begin{tabular}{@{}l|ccccc@{}}
\toprule
Method  & CC $ \uparrow $ & SCD $ \uparrow $ & PSNR $ \uparrow $ & N\textsuperscript{AB/F} $ \downarrow $ & NLPD $ \downarrow $ \\ \midrule
CUFD~\cite{CUFD}         & 0.463 & 1.221 & 61.70 & 0.027 & 0.5327 \\
DIVFusion~\cite{divfusion}    & 0.527 & 1.530 & 60.41 & 0.083 & 0.6152 \\
GAN-FM~\cite{GAN-FM}       & 0.548 & 1.700 & 62.29 & 0.028 & 0.5138 \\
NestFuse~\cite{NestFuse}     & 0.542 & 1.578 & \textcolor{blue}{62.76} & \textcolor{red}{0.009} & \textcolor{red}{0.4822} \\
STDFusionNet~\cite{STDFusionNet} & 0.459 & 1.179 & 62.40 & 0.028 & 0.5047 \\
SeAFusion~\cite{SeAfusion}    & 0.525 & 1.586 & 61.12 & 0.026 & 0.5098 \\
TarDAL~\cite{TAL}       & 0.481 & 1.443 & 58.98 & 0.185 & 0.6660 \\
MetaFusion~\cite{Metafusion}   & 0.555 & 1.682 & 62.42 & 0.147 & 0.6271 \\
PSFusion~\cite{PSfusion}     & \textcolor{blue}{0.559} & \textcolor{red}{1.832} & 60.80 & 0.086 & 0.5927 \\
MaeFuse      & \textcolor{red}{0.571} & \textcolor{blue}{1.750} & \textcolor{red}{63.66} & \textcolor{blue}{0.021} & \textcolor{blue}{0.5028} \\ \bottomrule
\end{tabular}
\label{tab: m3fd}
\end{table}

\textbf{Quantitative comparison.} We also report in Tables~\ref{tab: road},~\ref{tab: tno}, and~\ref{tab: m3fd} the quantitative comparison results of our method against nine other fusion methods across three datasets (RoadScene, TNO, M3FD). We used five objective metrics, namely CC, SCD, PSNR, Nabf, and NLPD, which are common evaluation indices. Among them, CC measures the linear correlation between two images, used to assess the similarity between images. A high CC value indicates a high degree of similarity. SCD is used to measure the similarity in the frequency domain between two images, applicable for evaluating the preservation of color or spectral information. PSNR is an indicator of image reconstruction quality, commonly used to assess the effects of image compression or restoration. A high PSNR usually indicates lower errors. The N\textsuperscript{AB/F} metric aims to quantify the number of artifacts in an image, used to assess the errors and anomalies introduced during image processing. A lower value indicates better performance. NLPD is used to evaluate the difference in visual quality between two images, especially in the details of multi-scale structures. And a lower value indicates better performance. It is clearly evident that our method demonstrates superiority in these statistical metrics.

\subsection{Ablation Study}

\begin{figure}[h] 
  \centering 
  \includegraphics[width=\columnwidth]{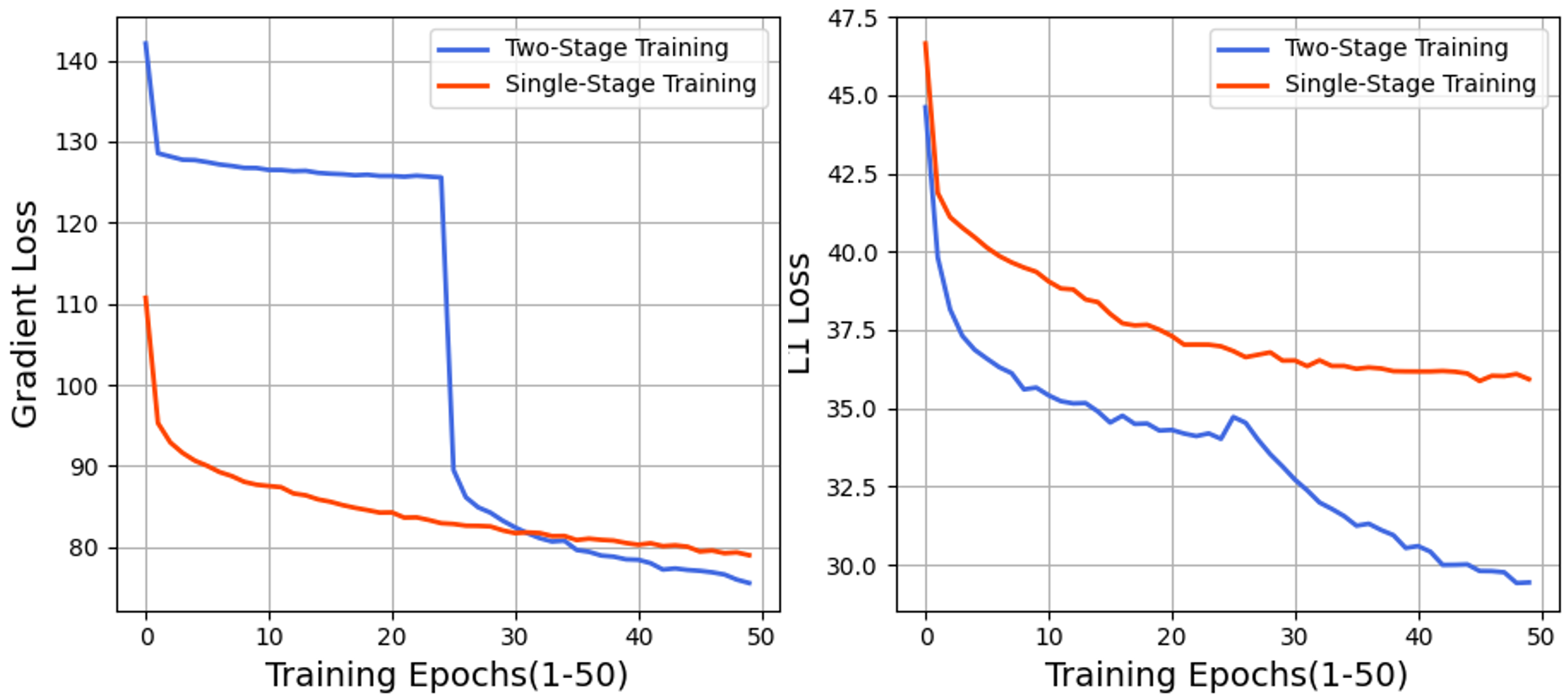} 
  \caption{The training process for two-stage training. The blue line represents two-stage training, with the first 25 iterations as the first stage and the latter 25 iterations as the second stage. The orange line represents training without using the two-stage approach. The left graph shows the gradient loss in fusion, while the right graph shows the $\mathcal{L}_1$ loss.}
  \vspace{-5pt}
  \label{fig:loss}
\end{figure}

\begin{figure}[h!] 
  \centering 
  \includegraphics[width=\columnwidth]{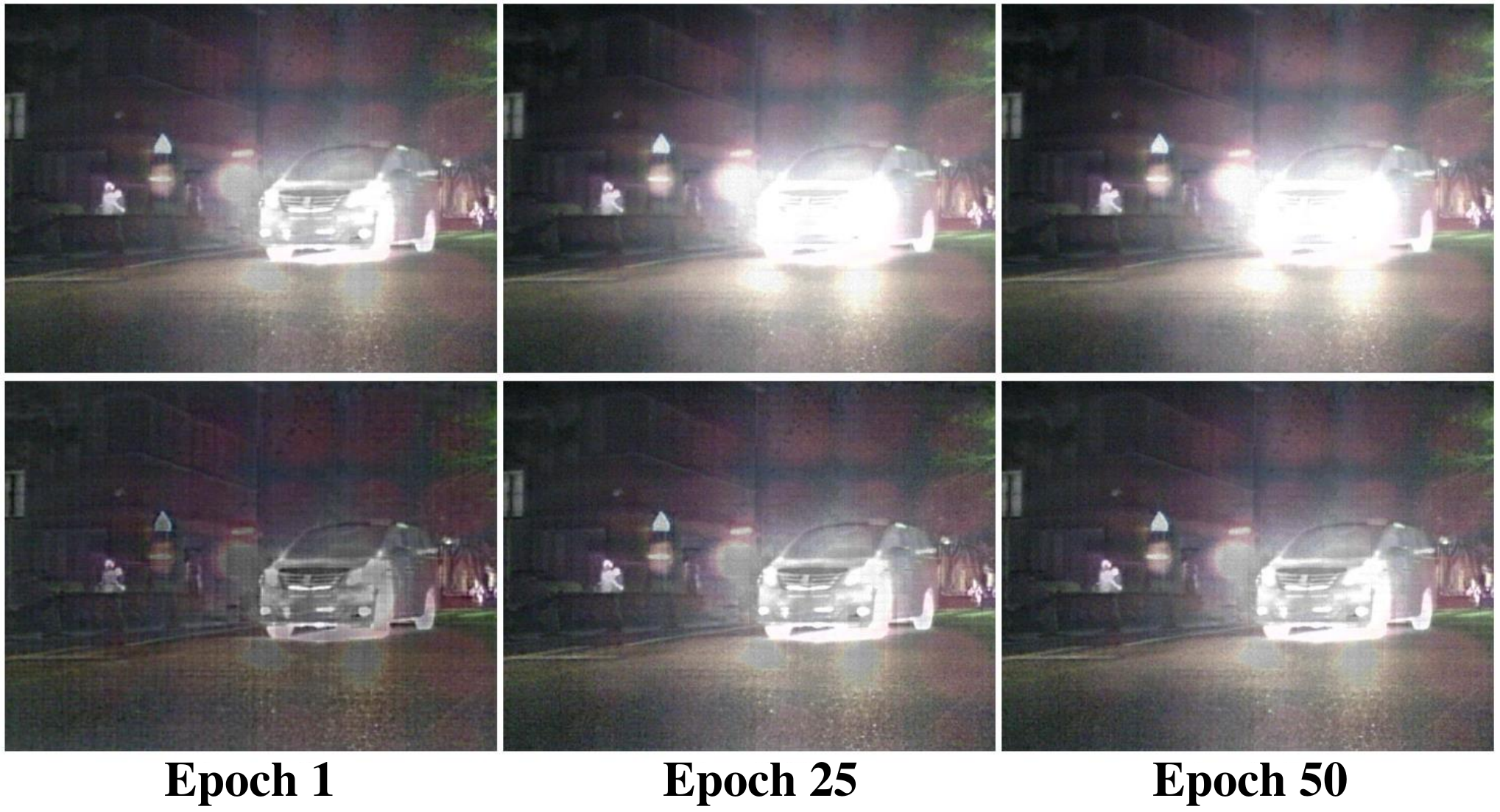} 
  \caption{Visualization results of hierarchical training ablation experiment according to different epoches. The first and second row shows the results with activated CFM weights and locked CFM weights, respectively.}
  \vspace{-15pt}
  \label{fig:step}
\end{figure}

\begin{figure*}[h!] 
  \centering 
  \includegraphics[width=\textwidth]{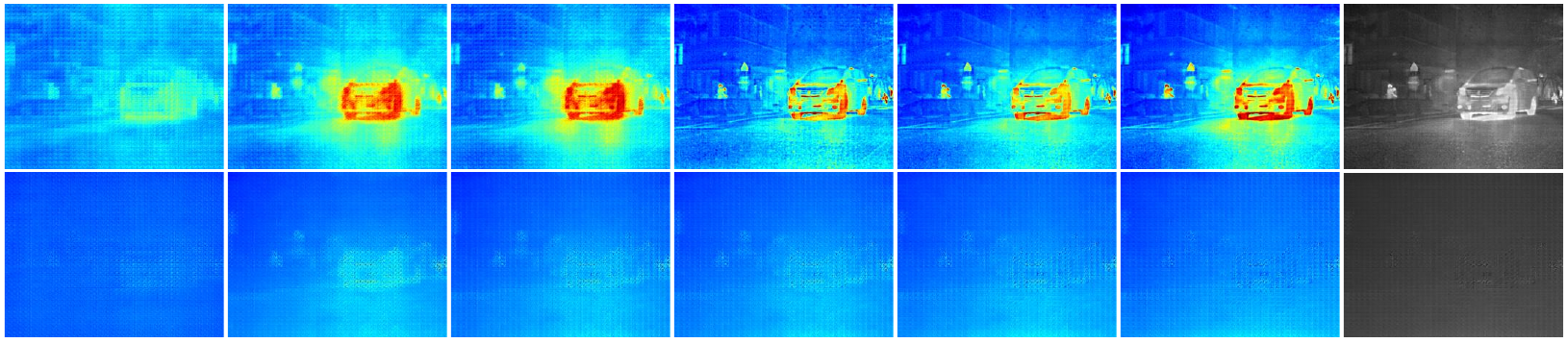} 
  \caption{Visualization results of two-stage training ablation studies. From left to right, the number of training iterations gradually increases. The first row shows the results using two-stage training; the first three images are from the first stage of training, and the next three images are from the second stage. The second row shows the results without using two-stage training, where each image corresponds to the same number of iterations as the images in the first row.}
  \vspace{-8pt}
  \label{fig:ab}
\end{figure*}

\begin{figure*}[h!] 
  \centering 
  \includegraphics[width=\textwidth]{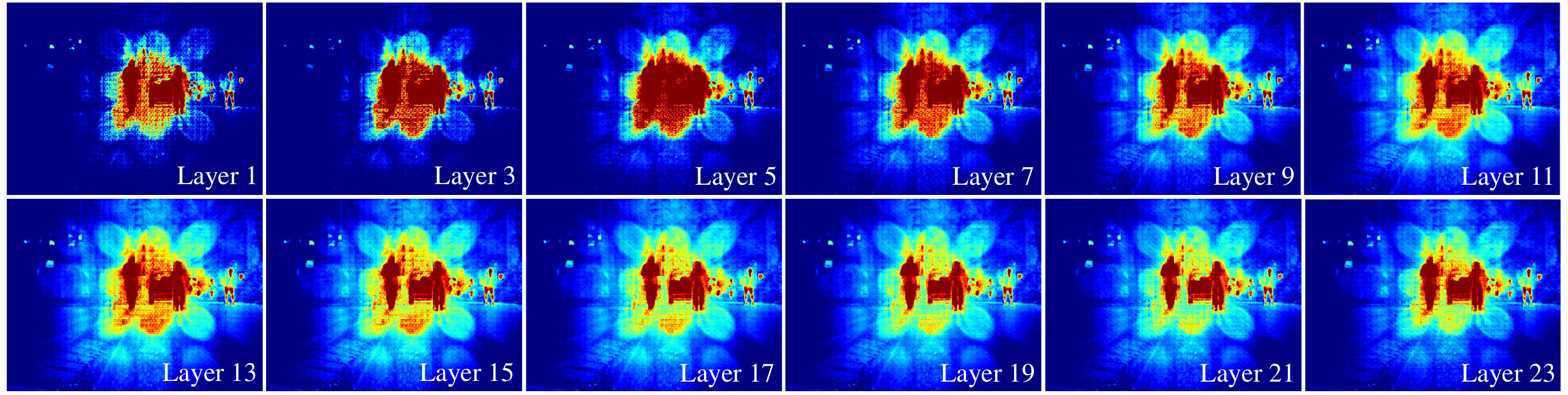} 
  \caption{Visualization results of the average fusion of feature vectors from different layers of the two modalities.}
  \vspace{-8pt}
  \label{fig:abh2}
\end{figure*}

To demonstrate the effectiveness of our guided training strategy, we conducted ablation experiments on both two-stage training and hierarchical training strategies. First, for the two-stage training, we trained a group of fusion layers directly with the fusion loss function for 50 iterations, while another group of fusion layers underwent domain alignment training for 25 iterations, followed by another 25 iterations of training with the same fusion loss function. In these two experiments, the hyperparameters used for the fusion loss functions are consistent with those in the original experiments. From the qualitative results, we observed that the data not undergoing two-stage training gradually failed to express image information, falling into local traps. In contrast, data with two-stage training rapidly aligned feature domains in the first 25 iterations, then refined image details with the fusion loss function in the latter 25 iterations, thus progressively eliminating visual issues caused by the ViT block effect. From a quantitative standpoint, the loss results without two-stage training gradually stabilized, indicating a local optimum trap. Meanwhile, data with two-stage training initially did not move towards the best fusion loss result but rapidly converged after domain alignment and achieved better outcomes, also showing a continuous downward trend. These points prove the effectiveness of our training method. The quantitative results are shown in Fig.~\ref{fig:loss}, while the qualitative results are presented in Fig.~\ref{fig:ab}.

To demonstrate the effectiveness of hierarchical training, we conducted an experiment using the same fusion training function. In this experiment, one group had the weights of the CFM activated, while the other group had the CFM weights locked. As shown in the results, the training with locked weights better maintained the original input features. In contrast, training with activated weights tended to cause the loss of original learning parameters, leading to overfitting of the loss function in the image results. Therefore, hierarchical training is beneficial for learning features content layer by layer. Qualitative results are shown in Fig.~\ref{fig:step}.

\begin{figure}[h] 
  \centering 
  \includegraphics[width=\columnwidth]{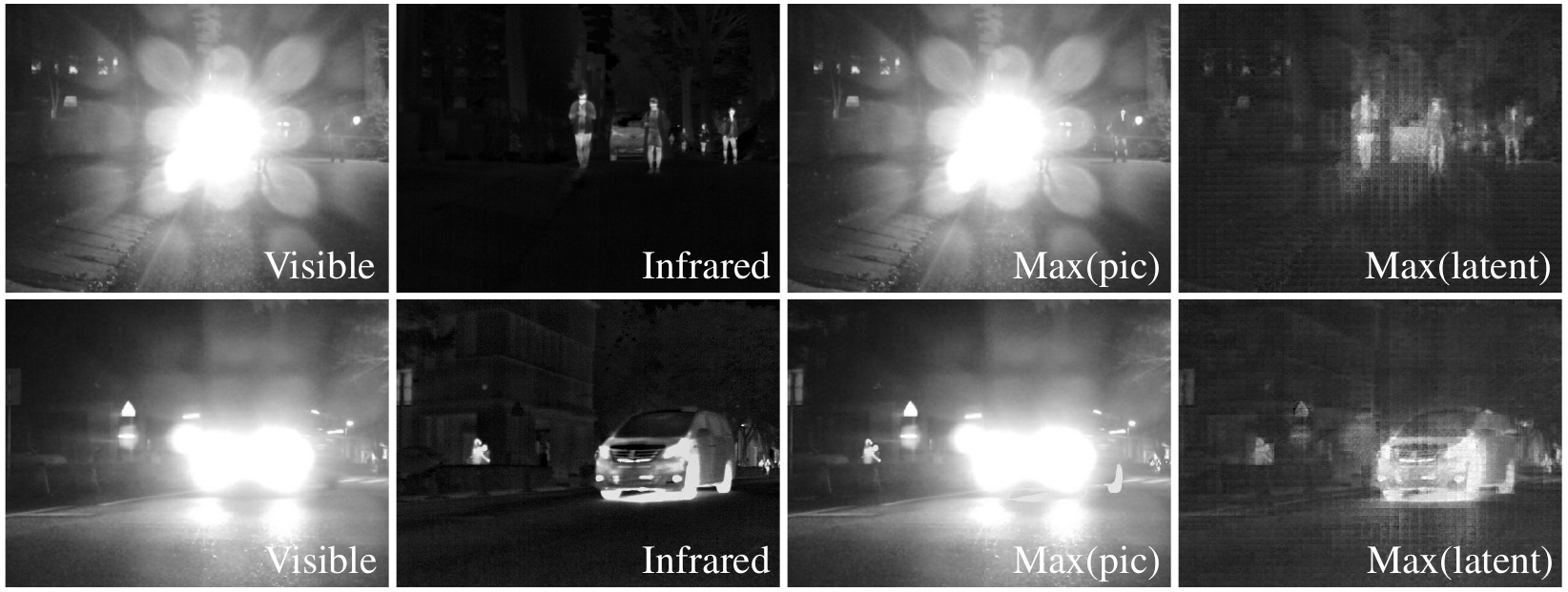} 
  \caption{The first and second rows show the `00037N' and `00909N' scenes from the MSRS dataset. The first column contains visible images, the second column contains infrared images, the third column shows the maximum value fusion in the image domain, and the fourth column shows the maximum value fusion in the feature domain.}
  \vspace{-10pt}
  \label{fig:abh1}
\end{figure}

\begin{figure*}[ht] 
  \centering 
  \includegraphics[width=0.98\textwidth]{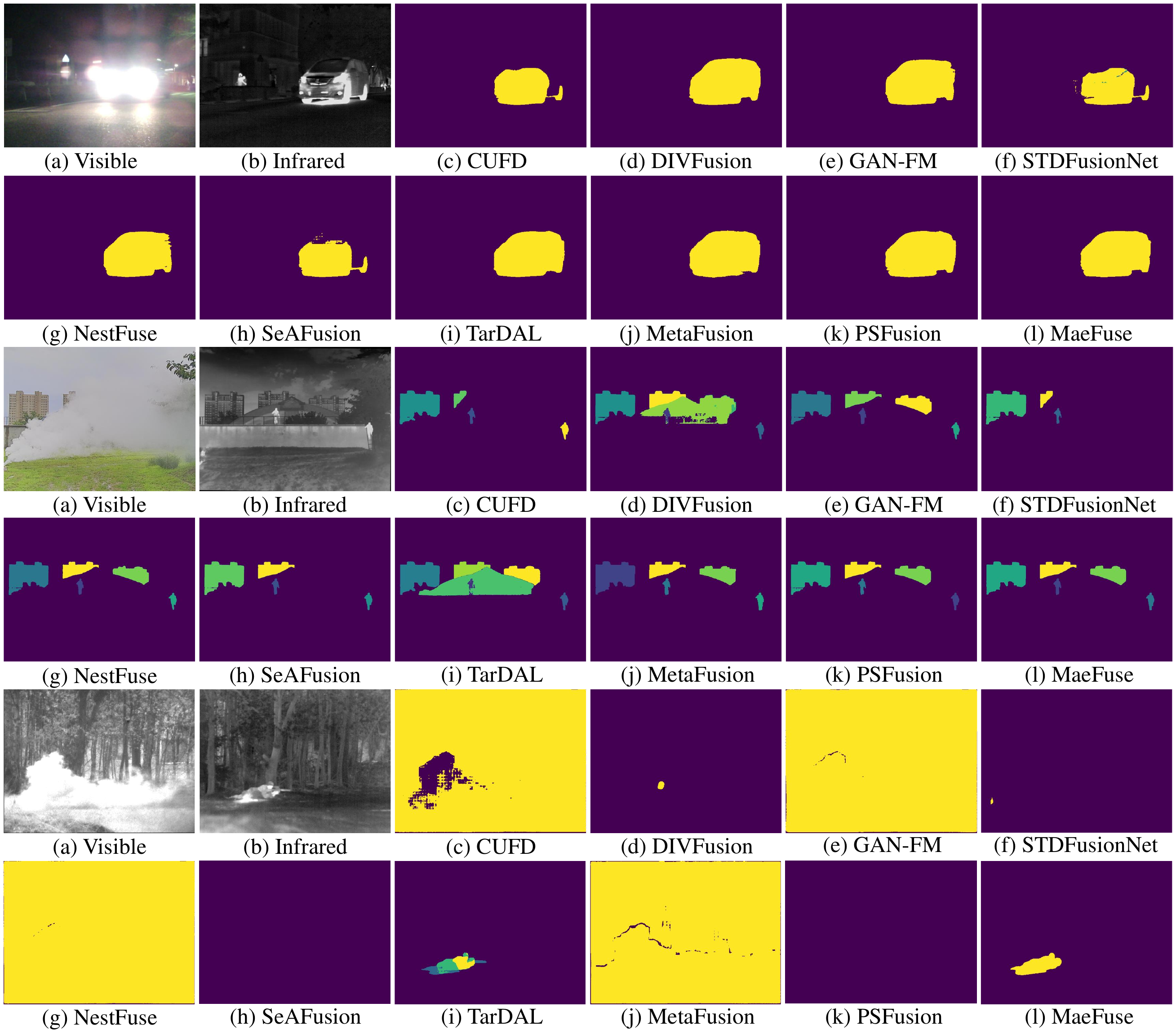} 
  \caption{The results of the qualitative analysis of the segmentation tasks, from top to bottom, every two rows form a comparison group, respectively. The prompts used for Grounded SAM of the first two rows, the second two rows, and the last two rows are `Car.', `Person. Building.', and `Soldier.', respectively.  
  }
  \vspace{-10pt}
  \label{fig:ab_3}
\end{figure*}

Since existing methods often utilize downstream tasks to assist with fusion training, our network is able to learn high-level semantic information. However, we have achieved the same or even better results without using downstream tasks. Here, I demonstrate through several ablation experiments that the feature vectors output by our MAE's pretrained encoder indeed possess high-level semantic information. In Fig.~\ref{fig:abh1}, we directly perform maximum value fusion of images from two modalities in the pixel domain, as well as in the feature domain. We find that performing maximum value fusion in the feature space does not result in overexposure, indicating that our feature vectors contain high-level semantic information and have greater weight. In Fig.~\ref{fig:abh2}, we visualize the results of averaging the features output from different layers of the MAE encoder. We find that as the number of layers increases, the high-level semantic information in the encoder becomes more apparent, showing that the MAE encoder can indeed capture sufficient high-level semantic information.

\subsection{Generalization Ability}

To verify that our fused images indeed retain sufficient high-level semantic information and are beneficial for downstream tasks, we conducted the following experiments for testing: semantic segmentation with prompts using Grounded SAM~\cite{groundedsam}, which first uses prompt words to identify the detection target, then uses SAM~\cite{sam} for semantic segmentation of the corresponding target. We used the most common segmentation methods to perform the segmentation directly and then compared the results. This allows us to fairly demonstrate whether the semantic information of the fused images can be easily recognized by existing segmentation models. If good results are obtained, it indicates that the information is sufficient and can be easily perceived by downstream tasks.




Existing segmentation techniques can perform well under various complex conditions, as shown in Fig.~\ref{fig:ab_3}:a and Fig.~\ref{fig:ab_3}:b, where the segmentation results are better than the corresponding visual results in Fig.~\ref{fig:ex-msrs2} and Fig.~\ref{fig:ex-m3fd}. Therefore, if it performs well visually, the downstream tasks will also perform well, which is the core viewpoint of the paper PSFusion~\cite{PSfusion}. Regarding the details in Fig.~\ref{fig:ab_3}:a, we can observe that our results do not have segmentation edge errors at the bottom or rear of the car. Similarly, in Fig.~\ref{fig:ab_3}:b, we can also effectively segment the three distant apartment buildings.

The comparison image in Fig.~\ref{fig:ab_3}:c is from the TNO~\cite{Toet2012ProgressIC} dataset, which does not have segmentation labels. Therefore, all works that use downstream tasks for auxiliary training have not seen these images. In the visual results, we can see many images that fail to highlight the soldier from the fog. As a result, many outcomes either do not generate a mask or result in erroneous full-screen segmentation. It is worth mentioning that for TarDAL~\cite{TAL}, PSFusion~\cite{PSfusion} can visually detect the soldier, but the segmentation results show multiple segmentations or missed detections, which we believe indicate poor generalization capabilities. The high-level semantic information in the fused results has a domain gap with that of general natural images, making it difficult to segment the desired results using the powerful SAM. However, our mask successfully obtains the corresponding segmentation results, indicating that our results are beneficial for downstream tasks and possess good generalization capabilities.

\section{Conclusion}\label{sec5}

In this paper, we explore the use of feature vectors obtained from the pretrained MAE encoder for image fusion. Using this method, we can provide sufficient high-level semantic information for the fusion results, and it is not necessary to use downstream tasks for auxiliary training. This simplifies the fusion network structure and training method. Additionally, we propose a two-stage training strategy that allows our fusion network's feature space to quickly align with the feature space represented by the original MAE encoder, solving the issues of limited training data and falling into local optima. Experiments show that using MaeFuse can match or even surpass fusion methods driven by semantic segmentation tasks. Employing Omni features for fusion can become the next direction in fusion work.

Using pretrained networks for fusion, we need to focus on whether the network can simultaneously capture high-level and low-level visual information. It is worth noting that recently, some~\cite{hsfusion} have also used CycleGAN~\cite{cyclegan} for fusion with good results, because CycleGAN, when performing domain transformations, needs to acquire both high-level and low-level visual information. From this perspective, the idea of using Omni features for fusion has significant room for development.

\ifCLASSOPTIONcaptionsoff
  \newpage
\fi

{
\bibliographystyle{IEEEtran}
\bibliography{sn-bibliography}
}

\end{document}